\documentclass[a4paper,fleqn]{cas-sc}

\usepackage[authoryear]{natbib}
\usepackage{tabularx}
\usepackage{booktabs}
\usepackage{array}
\usepackage{subcaption}
\usepackage{hyperref}
\hypersetup
{
 unicode=false,     % non-Latin characters in Acrobat's bookmarks
 pdftoolbar=true,    % show Acrobat's toolbar?
 pdfmenubar=true,    % show Acrobat's menu?
 pdffitwindow=false,   % window fit to page when opened
 pdfstartview={FitH},  % fits the width of the page to the window
 pdftitle={My title},  % title
 pdfauthor={Author},   % author
 pdfsubject={Subject},  % subject of the document
 pdfcreator={Creator},  % creator of the document
 pdfproducer={Producer}, % producer of the document
 pdfkeywords={keywords}, % list of keywords
 pdfnewwindow=true,   % links in new window
 colorlinks=true,    % false: boxed links; true: colored links
 linkcolor=red,     % color of internal links
 citecolor=blue,    % color of links to bibliography
 filecolor=magenta,   % color of file links
 urlcolor=cyan      % color of external links
}

\def\tsc#1{\csdef{#1}{\textsc{\lowercase{#1}}\xspace}}
\tsc{WGM}
\tsc{QE}
\begin{document}
\let\WriteBookmarks\relax
\def\floatpagepagefraction{1}
\def\textpagefraction{.001}

% Short title
\shorttitle{}    

% Short author
\shortauthors{}  

% Main title of the paper
\title [mode = title]{VISTA-DZ: Visual Semantic Trajectory Adaptation for Personalized Dilemma Zone Prediction}  

% Title footnote mark
% eg: \tnotemark[1]
% \tnotemark[1] 

% Title footnote 1.
% eg: \tnotetext[1]{Title footnote text}
% \tnotetext[1]{} 

% First author
%
% Options: Use if required
% eg: \author[1,3]{Author Name}[type=editor,
%       style=chinese,
%       auid=000,
%       bioid=1,
%       prefix=Sir,
%       orcid=0000-0000-0000-0000,
%       facebook=<facebook id>,
%       twitter=<twitter id>,
%       linkedin=<linkedin id>,
%       gplus=<gplus id>]

\author[1,2]{Chuheng Wei}[orcid=0000-0002-0747-9398]
\cormark[1]
\ead{chuheng.wei@email.ucr.edu}

% URL of the first author
% \ead[url]{}

% Credit authorship
% eg: \credit{Conceptualization of this study, Methodology, Software}
% \credit{}

% Address/affiliation

\author[2]{Ziye Qin}[orcid=0000-0001-5785-5331]
\ead{qinziye@hotmail}

\author[1]{Ziran Wang}[orcid=0000-0003-2702-7150]
\ead{ziran@ieee.org}

\author[2]{Guoyuan Wu}[orcid=0000-0001-6707-6366]
\ead{guoyuan.wu@ucr.edu}

\affiliation[1]{organization={Purdue University},
            addressline={550 Stadium Mall Drive},
            city={West Lafayette},
            postcode={47907}, 
            state={Indiana},
            country={USA}}

\affiliation[2]{organization={University of California, Riverside},
            addressline={1084 Columbia Ave}, 
            city={Riverside},
            postcode={92507}, 
            state={California},
            country={USA}}
            
% Corresponding author text
\cortext[1]{Corresponding author}

% Footnote text
% \fntext[1]{Under Review}

% For a title note without a number/mark
\nonumnote{This manuscript is currently under review.}

% Here goes the abstract
\begin{abstract}
Driver decision making in the dilemma zone at signalized intersections is safety critical, as vehicles approaching a yellow signal must decide whether to stop or proceed within limited time and distance margins. Accurate prediction of both stop-go decisions and decision timing is important for adaptive signal control, advanced driver assistance systems, and human-centered intelligent transportation applications. However, dilemma zone behavior is strongly driver dependent. Similar approach trajectories may lead to different decisions across drivers because of differences in risk preference, braking habit, and decision threshold. Existing personalized models often rely on handcrafted scalar descriptors, which provide useful but limited summaries of individual behavior.

This paper proposes \textbf{VISTA-DZ} (\textbf{Vi}sual and \textbf{S}emantic \textbf{T}rajectory \textbf{A}daptation for personalized \textbf{D}ilemma \textbf{Z}one prediction), a semantic-profile-conditioned framework for personalized stop-go and decision-time prediction. Historical driver trajectories are first converted into structured visual representations and interpreted by a vision-language model to produce natural-language behavioral profiles. These profiles are encoded into dense semantic embeddings and used to condition a dual-output trajectory prediction network. The final VISTA-DZ model combines a bidirectional GRU trajectory encoder with driver-conditioned multi-head cross-attention and Feature-wise Linear Modulation (FiLM), allowing the semantic profile to guide both temporal evidence selection and feature-level trajectory adaptation.

We evaluate VISTA-DZ on the Simulator Dilemma Zone (SDZ) dataset and a newly collected Field Dilemma Zone (FDZ) dataset using random split, leave-one-driver-out (LODO), zero-shot sim-to-real, and combined-domain evaluation protocols. Results show that semantic driver profiling improves over trajectory-only and handcrafted personalization baselines, while the final VISTA-DZ architecture achieves the best in-domain simulation accuracy of 93.26\%. In the LODO evaluation, VISTA-DZ obtains 90.22\% mean accuracy across 20 held-out simulation drivers, indicating that profile-conditioned prediction can generalize to drivers whose trajectories are not used for training. Cross-domain experiments further show that zero-shot simulation-to-real transfer is feasible, and that combining simulation data with limited real-world data provides the most reliable real-world generalization. These findings demonstrate the value of semantic driver profiling for personalized dilemma zone prediction and highlight the importance of target-domain calibration for field deployment. 

To support reproducibility and future research, the code and the FDZ dataset will be made publicly available upon acceptance.

\end{abstract}

% Use if graphical abstract is present
%\begin{graphicalabstract}
%\includegraphics{}
%\end{graphicalabstract}

% Research highlights
% \begin{highlights}
% \item A VLM-based personalized dilemma zone prediction framework is proposed to generate semantic driver profiles from historical trajectory visualizations.
% \item Two personalization strategies, feature concatenation and FiLM-based trajectory conditioning, are designed and systematically compared in the proposed framework.
% \item A controlled real-world dilemma zone driving dataset is collected and combined with simulator data for joint evaluation in virtual and real driving settings.
% \item Random split, leave-one-driver-out, and 3 $\times$ 3 cross-domain experiments are conducted to examine the applicability boundaries of personalized prediction
% \end{highlights}

%\nocite{*}

% Keywords
% Each keyword is seperated by \sep

\maketitle

% Main text
% \section{}\label{}

% \clearpage %%Remove this from your manuscript

% Figure
% \begin{figure}%[]
%   \centering
% %    \includegraphics{}
%     \caption{}\label{fig1}
% \end{figure}

% \begin{table}%[]
% \caption{}\label{tbl1}
% \begin{tabular*}{\tblwidth}{@{}LL@{}}
% \toprule
%   &  \\ % Table header row
% \midrule
%  & \\
%  & \\
%  & \\
%  & \\
% \bottomrule
% \end{tabular*}
% \end{table}

% Uncomment and use as the case may be
%\begin{theorem} 
%\end{theorem}

% Uncomment and use as the case may be
%\begin{lemma} 
%\end{lemma}

%% The Appendices part is started with the command \appendix;
%% appendix sections are then done as normal sections
%% \appendix

\section{Introduction}
\label{sec:introduction}

At signalized intersections, the onset of a yellow signal creates a short but safety-critical decision period for approaching drivers. Within the so-called dilemma zone (DZ), a vehicle may be too close to the stop line to stop comfortably, yet too far away to clear the intersection safely, making the stop-go decision uncertain and time-constrained~\citep {gazis1960problem,wei2024dilemma}. This phenomenon has long been associated with red-light running, abrupt braking, rear-end conflicts, and intersection crashes, and has therefore attracted sustained attention in traffic engineering and intelligent transportation systems~\citep{zegeer1977effectiveness,papaioannou2007driver,urbanik2007dilemma}. Reliable prediction in this setting requires more than identifying the eventual stop-go outcome. It is also important to estimate when the driver commits to that decision after yellow onset, since decision timing affects the available response window for adaptive signal control, advanced driver assistance systems (ADAS), and human-centered vehicle automation.

Recent trajectory-based learning methods have made it possible to model driver decisions directly from motion signals such as distance to the stop line, speed, and longitudinal acceleration. Recurrent neural networks and other deep sequence models can capture temporal patterns in vehicle motion and have shown promise for behavior prediction from driving trajectories~\citep{xu2022drone}. Nevertheless, dilemma zone behavior remains difficult to model with a single population-level predictor. Drivers differ in risk tolerance, braking strategy, preferred safety margin, and reaction timing when facing the same yellow-light scenario~\citep{papaioannou2007driver,wei2024dilemma}. A deceleration pattern that indicates a committed stop for one driver may still correspond to a delayed or aggressive go decision for another. Such heterogeneity motivates personalized dilemma zone modeling, especially for safety-critical applications that must anticipate individual intent rather than average population behavior~\citep{chauhan2022analysing}.

Most existing personalization strategies represent driver characteristics through a small number of handcrafted scalar descriptors, such as historical stop rate, mean approach speed, average decision distance, or average decision time. These descriptors are intuitive and easy to compute, but they compress within-driver behavior into low-dimensional summaries. As a result, they may discard important structure in how a driver approaches the stop line, adjusts speed, delays commitment, or changes braking intensity across different dilemma zone encounters~\citep{qin2025investigating}. Two drivers may have similar go probabilities, for example, while exhibiting very different behavioral mechanisms. One may behave consistently near the decision boundary, while another may alternate between early stopping and aggressive clearing. A more expressive driver representation is therefore needed, one that can capture behavioral style at a semantic level while remaining usable by downstream prediction models.

Vision-language models (VLMs) offer a promising way to build such representations. Instead of reducing historical trajectories to a few manually selected statistics, historical driving behavior can be transformed into structured trajectory visualizations and summarized in natural language. Language provides a compact but expressive medium for describing behavioral tendencies such as aggressiveness, braking style, speed consistency, and decision preference. Once encoded as dense embeddings, these semantic profiles can serve as driver-specific conditioning variables for trajectory prediction. This provides a bridge between low-level kinematic observations and higher-level descriptions of individual driving style.

In this paper, we propose \textbf{VISTA-DZ} (\textbf{Vi}sual and \textbf{S}emantic \textbf{T}rajectory \textbf{A}daptation for personalized \textbf{D}ilemma \textbf{Z}one prediction), a semantic-profile-conditioned framework for personalized dilemma zone prediction. For each driver, historical dilemma zone trajectories are converted into structured visual representations and interpreted by a VLM to generate a natural-language behavioral profile. The profile is then encoded into a dense semantic embedding and used to condition a dual-output prediction model. Given a three-second pre-yellow trajectory window, the model jointly predicts the stop-go decision and the decision time.

The final VISTA-DZ architecture uses a bidirectional GRU to encode the current approach trajectory and incorporates the semantic driver profile through two complementary mechanisms. First, the profile acts as a query in a driver-conditioned multi-head cross-attention module, allowing the model to attend to different temporal portions of the approach depending on the driver's behavioral tendency. Second, the attended trajectory representation is adapted through Feature-wise Linear Modulation (FiLM)~\citep{perez2018film}, which rescales and shifts trajectory features according to the semantic profile. In this way, VISTA-DZ does not simply append driver information to the trajectory feature. Instead, it uses the driver profile to guide both temporal evidence selection and feature-level interpretation.

To evaluate the proposed framework, we use both simulation and real-world dilemma zone data. The simulation dataset contains 961 trajectories from 20 drivers, while the controlled real-world dataset contains 172 trajectories from six drivers. The two datasets share the same trajectory representation and prediction targets, enabling evaluation across both within-domain and cross-domain settings. We consider random split evaluation on the simulation dataset, leave-one-driver-out (LODO) evaluation for unseen-driver generalization, zero-shot simulation-to-real transfer, and combined-domain training with both simulation and real-world data. This design allows us to examine not only whether semantic personalization improves prediction, but also when it generalizes across drivers and domains.

The main contributions of this paper are summarized as follows:
\begin{itemize}
    \item We propose VISTA-DZ, a semantic-profile-conditioned framework for personalized dilemma zone prediction that uses VLM-generated behavioral descriptions to represent individual driving style from historical trajectory evidence.

    \item We design a driver-conditioned trajectory adaptation architecture that combines bidirectional recurrent trajectory encoding, multi-head cross-attention, LayerNorm, and FiLM-based feature modulation for joint stop-go and decision-time prediction.

    \item We evaluate semantic driver profiling against trajectory-only prediction and handcrafted scalar personalization, showing that language-based profiles provide a stronger personalization signal and that modulation-based conditioning is more effective than direct feature concatenation.
    
    \item We collect a controlled real-world Field Dilemma Zone (FDZ) dataset of real-vehicle yellow-onset approaches, aligned with the simulator dataset for joint cross-domain evaluation, and will release the dataset and code upon acceptance.

    \item We conduct a comprehensive evaluation on both simulation and controlled real-world driving data, including random split, LODO, zero-shot sim-to-real, and combined-domain protocols, to assess in-domain performance, unseen-driver generalization, and cross-domain transfer.
\end{itemize}

\section{Related Work}
\subsection{Dilemma Zone Modeling and Stop-Go Prediction}

\begin{figure}
    \centering
    \includegraphics[width=0.6\linewidth]{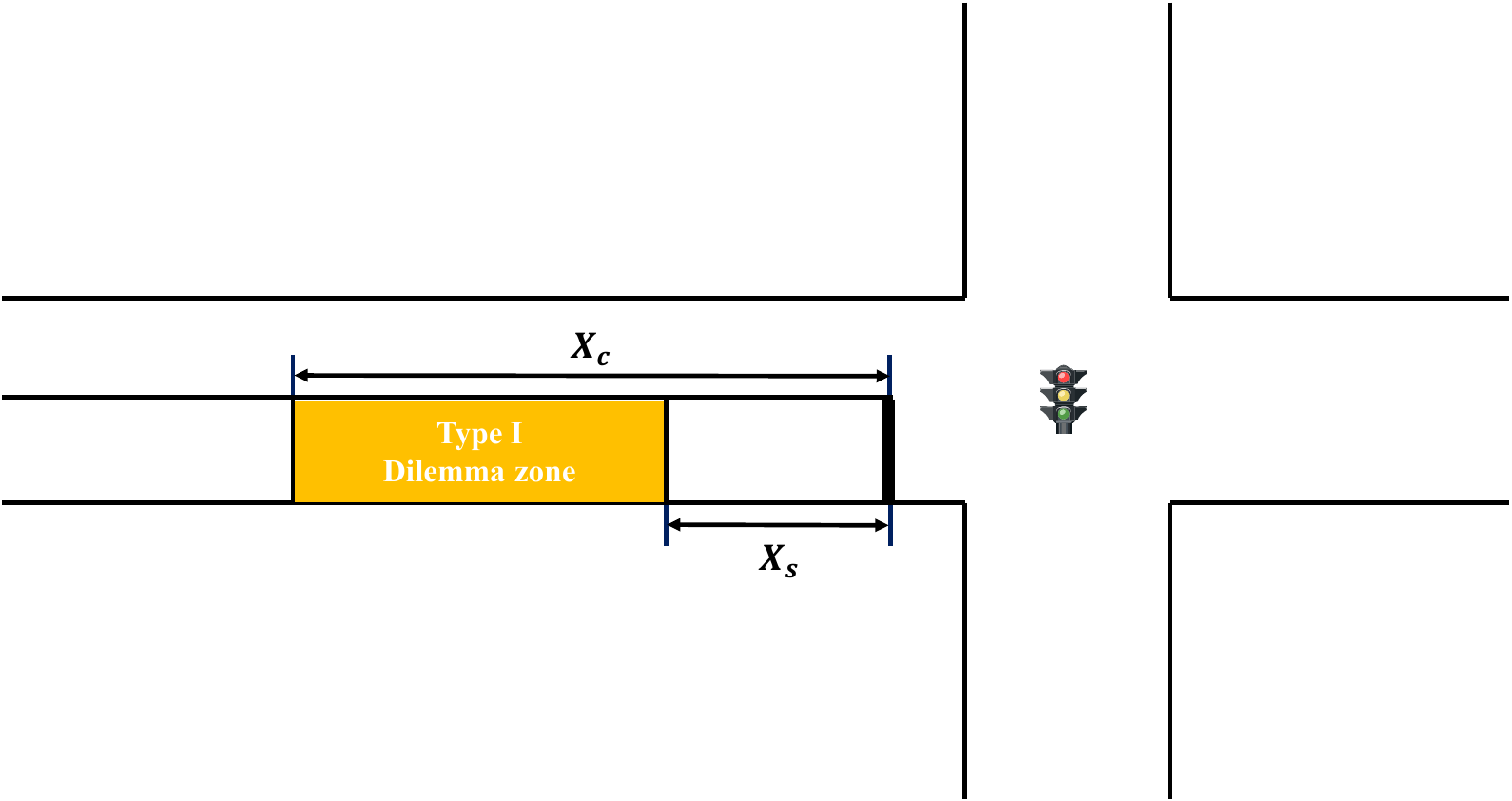}
    \caption{Type~I dilemma zone: a kinematic construct that arises when the maximum clearing distance $X_s$ is smaller than the minimum comfortable stopping distance $X_c$.}
    \label{fig:type1dz}
\end{figure}

\begin{figure}
    \centering
    \includegraphics[width=0.6\linewidth]{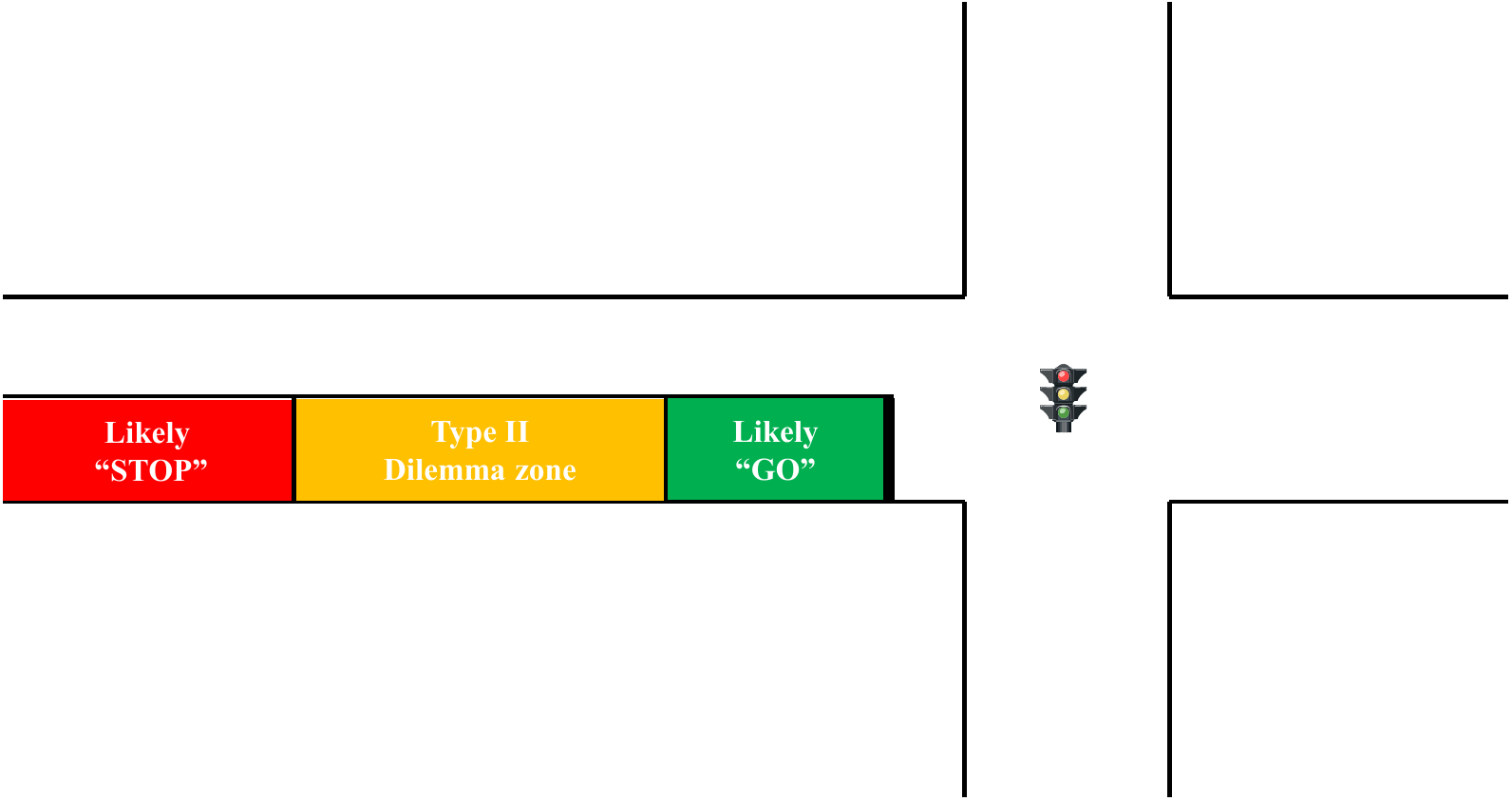}
    \caption{Type~II (indecision) dilemma zone: a behavior-based construct defined as the interval over which the empirical probability of stopping falls between 10\% and 90\%~\citep{qin2025investigating}.}
    \label{fig:type2dz}
\end{figure}

The dilemma zone has been studied for more than six decades, and two definitional traditions remain in use today. The kinematic definition introduced by \cite{gazis1960problem}, commonly referred to as the Type~I dilemma zone, identifies the region in which a vehicle can neither stop comfortably before the stop line nor clear the intersection within the remaining yellow interval. Its boundaries are derived from approach speed, comfortable deceleration, perception-reaction time, and intersection geometry, and depend on the relative ordering of two critical distances: the minimum distance $X_c$ needed to stop using a comfortable deceleration and the maximum distance $X_s$ from which the intersection can still be cleared during the yellow phase. When $X_s < X_c$, the interval between them forms the Type~I dilemma zone illustrated in Fig.~\ref{fig:type1dz}, within which the driver can neither stop nor proceed safely under standard kinematic assumptions. The probabilistic, behavior-based definition, proposed by \cite{zegeer1978green}, departs from this purely kinematic view and delineates the Type~II (or indecision) zone as the distance interval over which the observed stopping probability falls between 10\% and 90\%, as shown in Fig.~\ref{fig:type2dz}. Subsequent work has shifted the underlying boundary variable from spatial distance to time-to-stop-line: \cite{chang1985timing} formalized the travel-time view, \cite{bonneson2002intelligent} estimated the zone to span approximately 5.5-2.5~s of travel time, and \cite{rakha2008modeling} broadened this range to 5.5-1.5~s across age and gender groups. \cite{urbanik2007dilemma} further argued that practical detection-based protection must treat the dilemma zone as a behavioral construct rather than a purely kinematic one, since the kinematic envelope alone does not capture the variability of observed driver decisions.

\label{sec:rw_dz}

Beyond defining the zone itself, a substantial body of work has focused on predicting the binary stop-go outcome at yellow onset. Early models relied on parametric statistical formulations. \cite{papaioannou2007driver} applied binary logistic regression to relate stop-go decisions to approach speed, distance, and driver demographics at urban signalized intersections in Greece, while \cite{hurwitz2012fuzzy} introduced a fuzzy-logic representation that captured the gradual rather than abrupt transition between stopping and proceeding at high-speed approaches. As intersection sensor data became more accessible, machine learning methods were adopted to relax the linearity and distributional assumptions embedded in these earlier models. \cite{elhenawy2015modeling} benchmarked logistic regression against support vector machines, random forests, and adaptive boosting for yellow-light stop-run classification, reporting that tree-based ensembles offered the most stable performance, particularly under wet roadway conditions. ~\cite{jahangiri2016red} extended this direction by combining naturalistic driving and simulator data to predict red-light running violations several seconds in advance. More recently, research has moved toward trajectory-level representations that exploit the temporal structure of the approach rather than scalar snapshots at yellow onset. More recently, research has moved toward trajectory-level representations that exploit the temporal structure of the approach rather than scalar snapshots at yellow onset. Within this direction, our prior work has developed deep generative trajectory models that jointly encode signal state and multi-vehicle interaction for forecasting at signalized intersections~\citep{wei2024ki, wei2025eki}, surveyed the influential factors governing dilemma zone behavior~\citep{wei2024dilemma}, and empirically examined trajectory-driven stop-go dynamics under a controlled simulator study~\citep{qin2025investigating}, all of which suggest that snapshot-based descriptors are insufficient for characterizing the unfolding decision process.

Despite this progression from kinematic boundary models to data-driven trajectory predictors, the field has retained a largely homogeneous view of the driver. Most prediction models are trained and evaluated at the population level, and driver-related factors, when included at all, typically enter as coarse demographic groupings such as age or gender~\citep{rakha2008modeling,lavrenz2014modeling}. Empirical evidence, however, has repeatedly shown that stop-go thresholds, braking aggressiveness, and decision timing vary substantially across individuals in ways that cannot be explained by such grouping variables alone~\citep{papaioannou2007driver,chauhan2022analysing}. This mismatch between the heterogeneity observed in real drivers and the homogeneous assumptions embedded in existing models motivates an explicitly personalized formulation of dilemma zone prediction, in which individual behavioral style is treated as a first-class conditioning variable rather than a residual source of unmodeled noise.

\subsection{Personalized Driver Behavior Modeling}
\label{sec:rw_pers}

Efforts to model individual driving behavior fall along a spectrum of representational granularity. The coarsest approaches group drivers into a small number of stylistic categories, typically aggressive, moderate, and conservative, using clustering or supervised classification on summary statistics of speed, acceleration, jerk, and headway~\citep{liao2024review}. \cite{constantinescu2010driving} derived such categories from logged driving data using data mining techniques, and \cite{wang2017driving} later cast style classification as a semi-supervised SVM problem to reduce the labeling burden. \cite{johnson2011driving} showed that smartphone inertial signals alone are sufficient to recognize aggressive maneuvers, while \cite{murphey2009driver} treated jerk profiles as a compact indicator of style. 
In addition, \cite{wang2020driver} classified drivers into behavioral types with a $k$-nearest-neighbors scheme and fitted type-specific neural models to predict their speed-tracking errors, demonstrating the pipeline on both a game-engine simulator and a real vehicle. Broader surveys by \cite{sagberg2015review} and \cite{martinez2017driving} cover the definitions and recognition pipelines used across this literature.
 Categorical descriptors of this kind are interpretable but inherently coarse: two drivers in the same cluster can produce very different stop-go behavior depending on context. A finer but still scalar alternative is to retain the underlying summary statistics themselves, such as average approach speed, acceleration variance, time headway, and historical stop rate, and to attach them as driver descriptors for downstream prediction. This is the handcrafted personalization strategy that our experiments use as a baseline (HCF-Pers), and it inherits the same limitation: a small set of scalar moments captures what a driver does on average rather than how they behave trajectory by trajectory.

Learning-based representations aim to overcome this by extracting driver-specific vectors directly from raw behavioral data. \cite{hallac2016driver} showed that a single turn carries enough information to identify a driver from CAN-bus signals, and the subsequent Driver2Vec model~\citep{yang2021driver2vec} cast driver identification as a representation-learning problem and produced dense embeddings transferable to downstream behavioral tasks. Autoencoder and self-supervised variants extend this idea by learning latent style representations without explicit driver labels, while a parallel line of work has applied few-shot adaptation, building on the model-agnostic meta-learning framework of \cite{finn2017model}, to let trajectory predictors specialize to new drivers with limited data. These representations are richer than scalar statistics, but they share two practical drawbacks. First, they are typically optimized for an auxiliary objective such as driver identification or reconstruction, so the resulting vectors are not directly interpretable in terms of behavioral semantics. Second, training them well requires substantially more per-driver data than is usually available at signalized intersections, where a single driver may contribute only a handful of yellow onset events.

Once a driver representation is available, a separate design question concerns how it should interact with the per-trajectory features used for prediction. The simplest option is to concatenate the driver vector with the trajectory embedding, which treats personalization as an additional input. Feature-wise conditioning mechanisms such as the conditional instance normalization introduced by \cite{dumoulin2017learned} and the Feature-wise Linear Modulation (FiLM) mechanism of \cite{perez2018film} take a different route, using the conditioning vector to rescale and shift each feature channel so that the driver representation actively reshapes how trajectory evidence is interpreted. Cross-attention in the style of \cite{vaswani2017attention} addresses a complementary aspect of the same problem by letting the conditioning vector act as a query that selects which trajectory positions are most relevant for a given driver. The two families of mechanisms are not mutually exclusive: cross-attention controls which temporal evidence enters the prediction, while modulation controls how that evidence is interpreted once selected, and recent multimodal studies suggest that combining the two can be more expressive than either alone. In personalized driver behavior modeling, however, these mechanisms have rarely been studied side by side, and to our knowledge, their combined use has not been examined in the dilemma zone setting.

Two gaps, therefore, remain directly relevant to dilemma zone prediction: existing driver representations sit between handcrafted statistics that are interpretable but semantically thin and learned embeddings that are richer but opaque and data hungry, while the joint use of temporal selection and feature-level modulation that has proven effective in other multimodal tasks has not been examined here. Closing both motivates a representation that captures driving style at a semantic level, paired with a conditioning design that guides both which parts of the trajectory matter and how their features are interpreted for stop-go and decision-time prediction.

\subsection{Vision-Language Models for Driver Behavior Understanding}
\label{sec:rw_vlm}

Vision-language models (VLMs) and large language models (LLMs) have rapidly entered the autonomous driving literature~\cite{wei2026beyond}, supported by general-purpose backbones such as CLIP~\citep{radford2021learning}, BLIP-2~\citep{li2023blip}, LLaVA~\citep{liu2023visual}, and the Qwen-VL family~\citep{bai2023qwen}. Most application work has concentrated on scene-level understanding and end-to-end driving control. DriveGPT4~\citep{xu2024drivegpt4} grounded language in vehicle control to produce interpretable end-to-end driving outputs, and DriveLM~\citep{sima2024drivelm} formulated driving reasoning as graph visual question answering across perception, prediction, and planning. LMDrive~\citep{shao2024lmdrive} demonstrated closed-loop driving conditioned on natural-language instructions in CARLA, while DriveVLM~\citep{tian2024drivevlm} integrated chain-of-thought reasoning into the driving stack for complex urban scenes. From a planning perspective, GPT-Driver~\citep{mao2023gpt} casts motion planning as a language modeling task, and VLP~\citep{pan2024vlp} used vision-language priors to inject contextual knowledge into the planner. These works show that natural-language representations can capture aspects of driving that are difficult to articulate through purely numerical states, but their target is the scene and the policy rather than the individual driver.

A smaller line of work has begun to use language for driver-level rather than scene-level characterization. \cite{cui2024personalized} fine-tuned an LLM-based driving assistant on conversational personalization data and showed in field experiments that natural-language instructions can shape adaptive driving behavior. For longitudinal control, \cite{qin2025contextual} fine-tuned LLaMA-3 and Mistral on a synthetic dataset of driver preferences and contextual conditions, demonstrating that an LLM can identify and balance competing objectives such as efficiency, comfort, and safety when generating personalized cruise commands. These efforts indicate that language is a workable medium for expressing the qualitative attributes of how an individual driver behaves. The resulting representations, however, are consumed by the language model itself to produce control commands rather than being packaged as a portable representation for a separate predictor. Independently, sentence-level text encoders such as Sentence-BERT~\citep{reimers2019sentence} and MiniLM~\citep{wang2020minilm} have made it practical to convert natural-language descriptions into dense embeddings that integrate cleanly with downstream neural models, which makes language-derived driver descriptions usable as conditioning vectors rather than only as control outputs.

Despite this momentum, the use of VLMs for individual driver profiling from historical trajectory evidence remains rare. The dominant VLM-driving literature is occupied with scene or policy questions, and the language-personalized control work treats the language model as the controller rather than as a profile generator for a separate predictor. To our knowledge, no prior work has used a VLM to read a driver's historical trajectory evidence, summarize it as a semantic profile, and use that profile to condition downstream stop-go and decision time prediction in the dilemma zone. VISTA-DZ targets this gap.

\subsection{Generalization across Drivers and Driving Domains}
\label{sec:rw_generalization}

Evaluating whether a personalized model generalizes to an unseen driver is a long-standing concern in driver behavior research. LODO and subject-independent splits are commonly used for this purpose, and they typically reveal a gap between within-driver and across-driver performance that is most pronounced for personalization mechanisms tied to driver-specific signals~\citep{martinez2017driving}. Several adaptation strategies have been proposed to narrow this gap. Meta-learning approaches based on the model-agnostic framework of \cite{finn2017model} aim to produce models that can specialize to new drivers from only a few observed examples, and learned driver representations such as Driver2Vec~\citep{yang2021driver2vec} use auxiliary identification objectives to embed each driver into a vector space that transfers to downstream tasks. These methods reduce but do not eliminate the cold-start problem: a driver whose behavioral distribution lies outside the training population, or whose historical record is short or noisy, can still produce a representation that the predictor has little prior experience with.

Cross-domain transfer between simulated and real driving environments introduces a different set of challenges. Driving simulators such as CARLA~\citep{dosovitskiy2017carla} offer scale, controlled scenario design, and access to safety-critical situations that would be difficult or unsafe to collect on the road, and they have been used extensively to train end-to-end driving policies and behavior models. The trajectories these systems produce, however, differ from real vehicle data in sensor characteristics, vehicle dynamics, and driver behavior. Several strategies have been developed to close this gap. Domain randomization, introduced by \cite{tobin2017domain} for robotic perception and later extended to driving, exposes the model to enough simulator variation during training that the real environment lies within the training distribution. Adversarial domain adaptation methods such as DANN~\citep{ganin2015unsupervised} instead learn feature representations that are invariant to the source-target distinction. \cite{bewley2019learning} demonstrated that a driving policy trained entirely in simulation can be transferred to a real vehicle with appropriate adaptation, and \cite{mueller2018driving} showed that policy transfer is more robust when intermediate representations are made explicitly invariant to surface appearance. Beyond these adaptation methods, a simpler and frequently effective approach is to train jointly on simulated and real data, using real samples to calibrate features that the model would otherwise overfit to simulator-specific patterns.

Joint evaluation of unseen-driver generalization and sim-to-real transfer in personalized dilemma zone prediction remains largely absent. Existing LODO studies tend to focus on within-domain prediction, and sim-to-real work in driving has concentrated on perception or full driving policy rather than on stop-go and decision time prediction at signalized intersections. Whether a personalized model that performs well within a single simulation domain still works when the driver is unseen, when the data domain shifts to real vehicles, or when both shifts occur simultaneously has not been examined for dilemma zone behavior. The $3 \times 3$ cross-domain protocol, together with the LODO evaluation in this paper, is designed to fill this gap.

\section{Problem Formulation}

This study considers personalized driver decision prediction in the dilemma zone at signalized intersections. For each driving instance, the goal is to predict both the driver's stop-go decision after yellow onset and the time at which the decision is made. In addition to defining the prediction target itself, the problem must also account for two practical challenges: inter-driver heterogeneity and distribution shift across different evaluation settings. The former motivates personalized modeling, while the latter requires explicit analysis of generalization to unseen drivers and transfer between virtual and real driving environments.

\subsection{Input and Output Definitions}

Let $\mathbf{X}_i$ denote the trajectory sample associated with the $i$-th driving instance. Each sample consists of a fixed-length observation window collected after the yellow onset as the vehicle approaches the intersection:
\begin{equation}
\mathbf{X}_i = [\mathbf{x}_{i,1}, \mathbf{x}_{i,2}, \dots, \mathbf{x}_{i,T}] \in \mathbb{R}^{T \times C},
\end{equation}
where $T$ is the number of time steps and $C$ is the number of input channels. In this work, $T=93$ and $C=3$. Each time step is represented by
\begin{equation}
\mathbf{x}_{i,t} = [d_{i,t},\, v_{i,t},\, a_{i,t}],
\end{equation}
where $d_{i,t}$ denotes the distance to the stop line, $v_{i,t}$ denotes the vehicle speed, and $a_{i,t}$ denotes the longitudinal acceleration at time step $t$.

For each trajectory sample, two prediction targets are defined. The first is a binary stop-go decision label,
\begin{equation}
y_i^{cls} \in \{0,1\},
\end{equation}
where $y_i^{cls}=0$ indicates that the driver stops and $y_i^{cls}=1$ indicates that the driver proceeds through the intersection. The second is a continuous decision-time label,
\begin{equation}
y_i^{reg} \in \mathbb{R},
\end{equation}
which represents the elapsed time between the yellow onset and the driver's commitment to the stop-go decision.

The learning task is therefore formulated as a joint prediction problem that maps the observed trajectory window to both a classification output and a regression output:
\begin{equation}
f(\mathbf{X}_i) \rightarrow (\hat{y}_i^{cls},\, \hat{y}_i^{reg}),
\end{equation}
where $\hat{y}_i^{cls}$ and $\hat{y}_i^{reg}$ denote the predicted stop-go decision and predicted decision time, respectively. This formulation captures both the discrete behavioral outcome and the temporal dynamics of the decision process.

\subsection{Personalized Prediction Setting}

Dilemma zone behavior varies substantially across drivers. Even under similar approach conditions, drivers may differ in braking style, response timing, and willingness to proceed through the intersection. To account for this heterogeneity, each sample is associated with a driver identity $u_i \in \mathcal{U}$, where $\mathcal{U}$ denotes the set of all drivers in the dataset.

In the personalized setting, each driver $u_i$ is further associated with a driver-specific profile vector
\begin{equation}
\mathbf{p}_{u_i} \in \mathbb{R}^{m},
\end{equation}
where $m$ denotes the profile dimension. This vector represents behavioral information derived from the driver's historical data and is used to condition the prediction model. The personalized prediction task can then be written as
\begin{equation}
f(\mathbf{X}_i,\, \mathbf{p}_{u_i}) \rightarrow (\hat{y}_i^{cls},\, \hat{y}_i^{reg}).
\end{equation}

Under this formulation, the trajectory $\mathbf{X}_i$ captures the instantaneous motion pattern of the current approach, while $\mathbf{p}_{u_i}$ provides driver-level context that reflects persistent behavioral tendencies. In other words, the model is expected to interpret the same trajectory pattern differently when it is associated with different driver profiles. This setting is more appropriate for dilemma zone prediction than a purely global formulation, because the boundary between stopping and proceeding is not determined by kinematics alone, but also by individual driving style.

In this paper, the driver profile is treated as a general personalization variable at the formulation level. Its specific construction is introduced in the methodology section, where both conventional handcrafted driver descriptors and VLM-generated semantic driver profiles are considered.

\subsection{Generalization and Cross-Domain Settings}

Beyond prediction accuracy under standard data splits, this work also examines how personalized models generalize under more challenging conditions. Two types of generalization are considered: cross-driver generalization and cross-domain generalization.

Cross-driver generalization addresses whether a model trained on one set of drivers can maintain performance when applied to a driver whose behavior has not been observed during training. In the standard random-split setting, the training and test sets may contain different trajectory samples from the same driver. This setting evaluates how effectively the model can exploit driver-specific information when the driver is represented in the training data. In contrast, the LODO setting enforces driver-level separation between training and testing. Let $\mathcal{U}_{train}$ and $\mathcal{U}_{test}$ denote the driver sets used for training and evaluation, respectively. Under LODO,
\begin{equation}
\mathcal{U}_{train} \cap \mathcal{U}_{test} = \emptyset.
\end{equation}
This setting is used to evaluate generalization to unseen drivers and to assess the robustness of personalization under a cold-start scenario.

Cross-domain generalization addresses transfer across different data sources. In this work, the data come from two domains: a simulation domain and a controlled real-world driving domain. Let $\mathcal{D}^{sim}$ denote the set of samples collected in the simulator and $\mathcal{D}^{real}$ denote the set of samples collected in real vehicle experiments under controlled field conditions. In addition, a combined domain is defined by
\begin{equation}
\mathcal{D}^{comb} = \mathcal{D}^{sim} \cup \mathcal{D}^{real}.
\end{equation}
The cross-domain setting considers training on one of these domains and evaluating on the same or another domain, thereby forming a $3 \times 3$ train-test evaluation matrix. This design makes it possible to examine in-domain performance, simulator-to-real transfer, real-to-simulator transfer, and the effect of joint training on both domains.

Taken together, the random split, LODO, and cross-domain settings provide a more complete view of personalized dilemma zone prediction than a single evaluation protocol. They make it possible to distinguish between three related but different questions: whether personalization improves prediction for observed drivers, whether it generalizes to unseen drivers, and whether it remains reliable across virtual and real driving environments.

\begin{figure}
    \centering
    \includegraphics[width=1\linewidth]{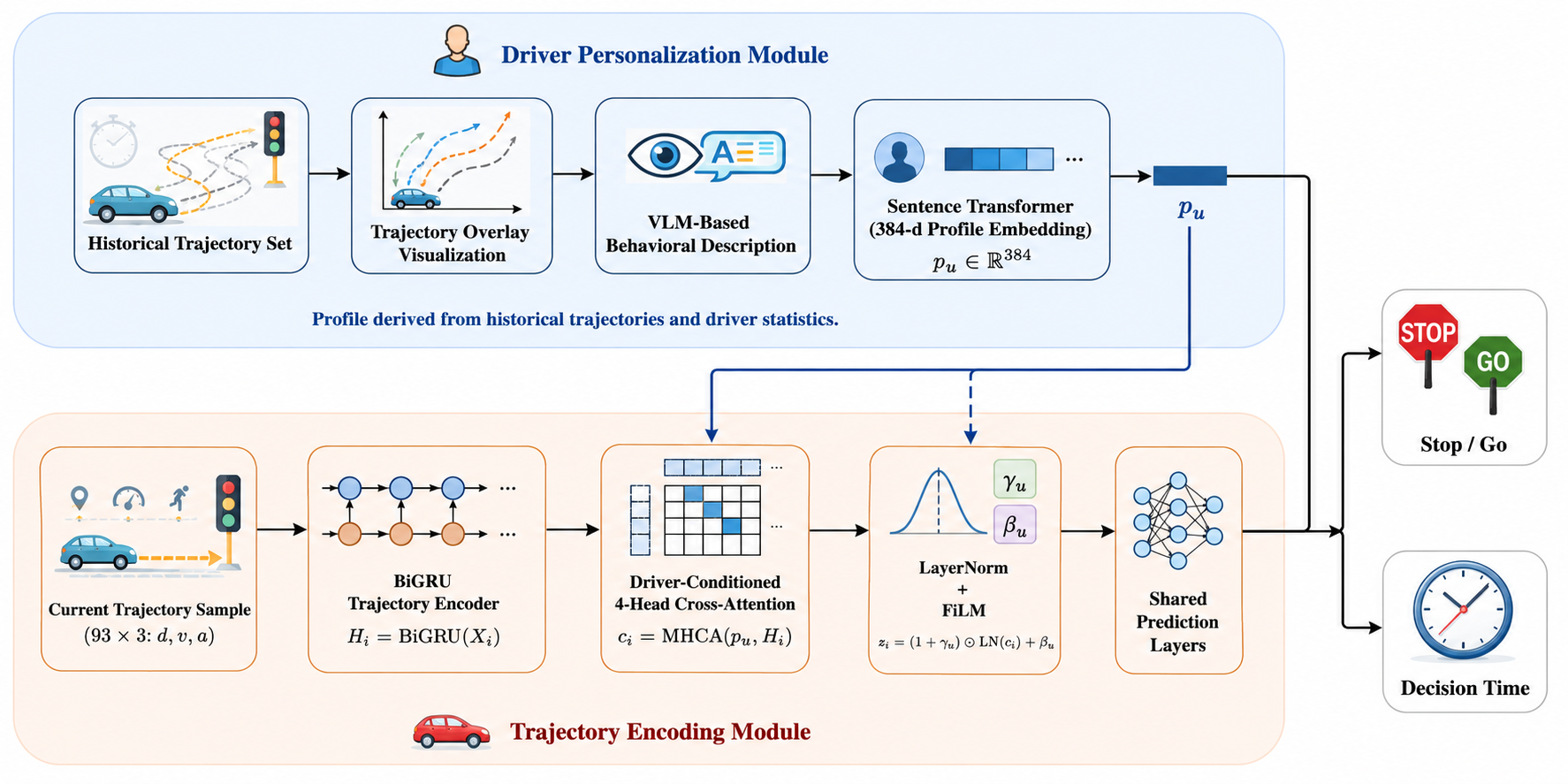}
        \caption{Overall framework of the proposed VISTA-DZ model. Historical dilemma zone trajectories are transformed into a semantic driver profile $\mathbf{p}_u$ through trajectory overlay visualization, VLM-based behavioral description, and sentence-transformer embedding. For the current trajectory sample $\mathbf{X}_i$, a BiGRU encoder extracts temporal hidden states, which are personalized by the driver profile through driver-conditioned multi-head cross-attention and FiLM-based feature modulation. The resulting adapted representation is passed to shared prediction layers to jointly predict the stop-go decision and decision time.}

    \label{fig:framework}
\end{figure}

\section{Methodology}
\label{sec:methodology}

This section presents the proposed \textbf{Visual Semantic Trajectory Adaptation for personalized Dilemma Zone prediction (VISTA-DZ)} framework. The goal is to predict both the stop-go decision and the decision time of a driver approaching a signalized intersection under yellow-signal onset. VISTA-DZ combines two sources of information: the short-term kinematic evidence from the current pre-yellow trajectory and the long-term behavioral tendency of the corresponding driver. The latter is represented by a semantic driver profile generated from historical dilemma zone trajectories and then used to personalize the trajectory encoder through driver-conditioned multi-head cross-attention and Feature-wise Linear Modulation (FiLM).

\subsection{Problem Formulation}
\label{subsec:problem_formulation}

For the $i$-th driving instance, let $u_i$ denote the driver identity. The model input is a fixed-length trajectory window observed before the yellow signal appears:
\begin{equation}
\mathbf{X}_i = \left[\mathbf{x}_{i,1}, \mathbf{x}_{i,2}, \dots, \mathbf{x}_{i,T}\right] \in \mathbb{R}^{T \times 3},
\end{equation}
where $T=93$ corresponds to approximately three seconds of pre-yellow motion sampled at around 31 Hz. Each timestep contains three kinematic channels:
\begin{equation}
\mathbf{x}_{i,t} = \left[d_{i,t}, v_{i,t}, a_{i,t}\right],
\end{equation}
where $d_{i,t}$ is the distance to the stop line, $v_{i,t}$ is the longitudinal speed, and $a_{i,t}$ is the longitudinal acceleration.

The prediction target contains two outputs. The first is a binary stop-go decision:
\begin{equation}
y_i^{sg} \in \{0,1\},
\end{equation}
where $0$ denotes stop and $1$ denotes go. The second is the decision time:
\begin{equation}
y_i^{dt} \in \mathbb{R},
\end{equation}
which measures the time elapsed from the yellow onset to the driver's decision commitment.

In addition to the current trajectory, VISTA-DZ uses a driver-specific semantic profile vector:
\begin{equation}
\mathbf{p}_{u_i} \in \mathbb{R}^{384},
\end{equation}
which summarizes the historical behavioral tendency of driver $u_i$. The overall prediction function is therefore written as
\begin{equation}
f\left(\mathbf{X}_i, \mathbf{p}_{u_i}\right) \rightarrow \left(\hat{o}_i^{sg}, \hat{y}_i^{dt}\right),
\end{equation}
where $\hat{o}_i^{sg}$ is the predicted stop-go logit and $\hat{y}_i^{dt}$ is the predicted decision time. At inference time, the stop-go class is obtained using a zero-logit threshold:
\begin{equation}
\hat{y}_i^{sg} = \mathbb{I}\left(\hat{o}_i^{sg} > 0\right),
\end{equation}
which is equivalent to thresholding the sigmoid probability at 0.5.

\subsection{Semantic Driver Profile Generation}
\label{subsec:profile_generation}

The central assumption of VISTA-DZ is that dilemma zone behavior is driver-dependent. For the same speed, distance, and acceleration pattern, different drivers may make different stop-go decisions due to differences in risk tolerance, braking style, and decision consistency. To represent these driver-level tendencies, VISTA-DZ constructs a semantic driver profile from historical dilemma zone trajectories.

For a driver $u \in \mathcal{U}$, let
\begin{equation}
\mathcal{H}_u = \left\{\mathbf{X}_u^{(1)}, \mathbf{X}_u^{(2)}, \dots, \mathbf{X}_u^{(N_u)}\right\}
\end{equation}
denote the set of historical trajectory samples associated with that driver. In addition to raw trajectories, a set of driver-level behavioral statistics is computed:
\begin{equation}
\mathbf{r}_u = \left[g_u, \bar{v}_u, \bar{d}_u, \bar{\tau}_u, \sigma_v^u, \delta_u^{dec}\right],
\end{equation}
where $g_u$ is the driver's go probability, $\bar{v}_u$ is the average speed at the decision point, $\bar{d}_u$ is the average distance to the stop line at the decision point, $\bar{\tau}_u$ is the average decision time, $\sigma_v^u$ measures speed consistency, and $\delta_u^{dec}$ denotes the maximum deceleration-related statistic extracted from the driver's historical trajectories.

\subsubsection{Trajectory Visualization and VLM Description}
\label{subsubsec:trajectory_visualization}

The historical trajectory set $\mathcal{H}_u$ is converted into a structured trajectory visualization:
\begin{equation}
\mathbf{V}_u = \mathcal{G}\left(\mathcal{H}_u\right),
\end{equation}
where $\mathcal{G}(\cdot)$ denotes the visualization operator. The visualization overlays repeated dilemma zone approaches from the same driver. Distance to the stop line and speed define the two primary axes, acceleration is represented through color, and the final stop/go outcome is distinguished using line style. This representation provides a compact visual summary of how a driver approaches the stop line, whether they maintain speed, decelerate early, or apply stronger braking closer to the intersection.

The trajectory visualization $\mathbf{V}_u$ and the structured statistics $\mathbf{r}_u$ are then provided to a vision-language model to produce a natural-language behavioral description:
\begin{equation}
\mathbf{s}_u = \Phi_{\mathrm{vlm}}\left(\mathbf{V}_u, \mathbf{r}_u\right),
\end{equation}
where $\Phi_{\mathrm{vlm}}(\cdot)$ denotes the VLM inference process and $\mathbf{s}_u$ is a textual description of driver $u$. In implementation, Qwen2.5-VL:7B is used to generate the driver description from the trajectory visualization and the structured prompt. The generated description summarizes behavioral dimensions such as decision tendency, aggressiveness, braking style, and speed consistency.

\begin{figure}
    \centering
    \includegraphics[width=1\linewidth]{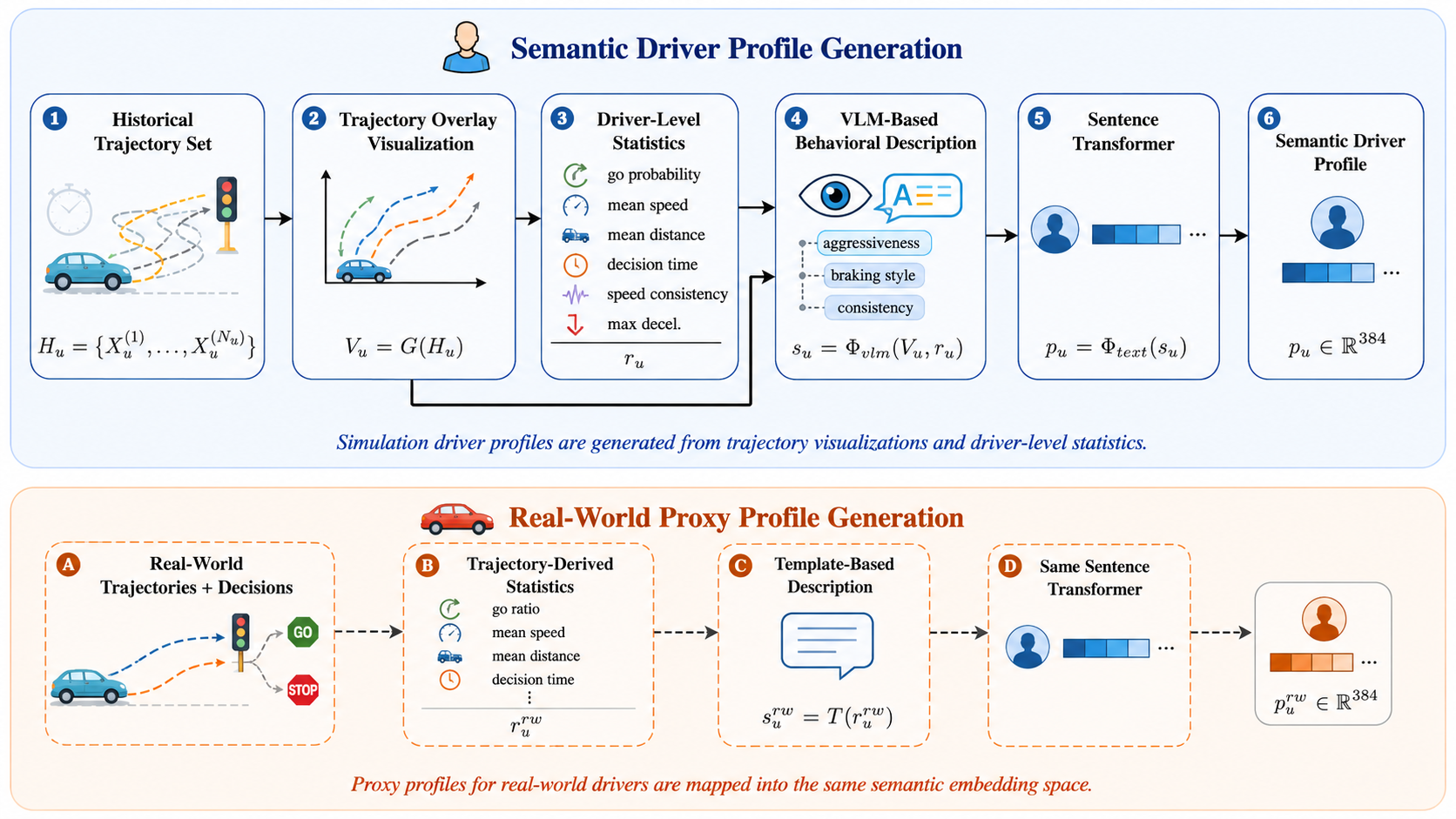}
    \caption{Generation of semantic driver profiles. Historical trajectory patterns and driver-level statistics are converted into a VLM-based behavioral description and then encoded into a 384-dimensional profile embedding $\mathbf{p}_u$. Real-world proxy profiles are generated through template-based descriptions and encoded using the same sentence transformer.}
    \label{fig:profile_generation}
\end{figure}

\subsubsection{Text Embedding}
\label{subsubsec:text_embedding}

The natural-language driver description is encoded into a dense vector using a sentence transformer:
\begin{equation}
\mathbf{p}_u = \Phi_{\mathrm{text}}\left(\mathbf{s}_u\right), \qquad \mathbf{p}_u \in \mathbb{R}^{384},
\end{equation}
where $\Phi_{\mathrm{text}}(\cdot)$ denotes the text encoder. In this work, all-MiniLM-L6-v2 is used as the text embedding model, yielding a 384-dimensional semantic profile for each driver. The resulting profile is treated as a fixed driver-level conditioning variable and is shared by all prediction samples associated with the same driver.

This profile generation process does not simply append manually selected statistics to the prediction model. Instead, it restructures historical trajectory patterns and driver-level statistics into an interpretable semantic representation. This design allows the subsequent prediction network to use driver-level behavioral context while still operating on the current trajectory window for sample-level decision prediction.

\subsubsection{Proxy Profiles for Real-World Drivers}
\label{subsubsec:proxy_profiles}

For real-world drivers, the same semantic profile format is required so that simulation and real-world data can be used within a unified model. However, real-world driver histories are typically smaller and noisier, and pre-existing personalized statistics may not be available in the same format as the simulation data. Therefore, for real-world drivers, VISTA-DZ constructs template-based proxy descriptions from trajectory-derived statistics.

For a real-world driver $u$, statistics such as go probability, mean speed at yellow onset, maximum deceleration, and speed consistency are computed directly from the available real-world trajectories and decisions. These statistics are inserted into a fixed natural-language template that follows the same linguistic structure as the VLM-generated simulation profiles. The resulting text is encoded using the same sentence transformer:
\begin{equation}
\mathbf{p}_u^{rw} = \Phi_{\mathrm{text}}\left(\mathcal{T}(\mathbf{r}_u^{rw})\right),
\end{equation}
where $\mathcal{T}(\cdot)$ denotes the template-based description generator and $\mathbf{r}_u^{rw}$ denotes the real-world driver statistics. Using the same text encoder ensures that simulation profiles and real-world proxy profiles lie in the same semantic embedding space.

In all experiments, the driver profile is treated as prior driver-level information. It is not updated by the prediction network during training or inference. The current trajectory sample $\mathbf{X}_i$ provides the short-term evidence for the specific decision event, while $\mathbf{p}_{u_i}$ provides the long-term behavioral context of the corresponding driver.

\subsection{VISTA-DZ Prediction Network}
\label{subsec:prediction_network}

Given the current trajectory $\mathbf{X}_i$ and the semantic driver profile $\mathbf{p}_{u_i}$, the VISTA-DZ prediction network first encodes the trajectory using a Bidirectional GRU (BiGRU), and then uses the driver profile to personalize the encoded temporal representation. Specifically, the driver profile is integrated into the prediction network through two consecutive conditioning mechanisms. First, it serves as the query in a driver-conditioned multi-head cross-attention module to select driver-relevant temporal evidence from the BiGRU hidden states. Second, it generates FiLM parameters to modulate the attended trajectory context in a feature-wise manner. The resulting personalized representation is then passed to shared prediction layers for joint stop-go classification and decision-time regression.

\begin{figure}
    \centering
    \includegraphics[width=0.7\linewidth]{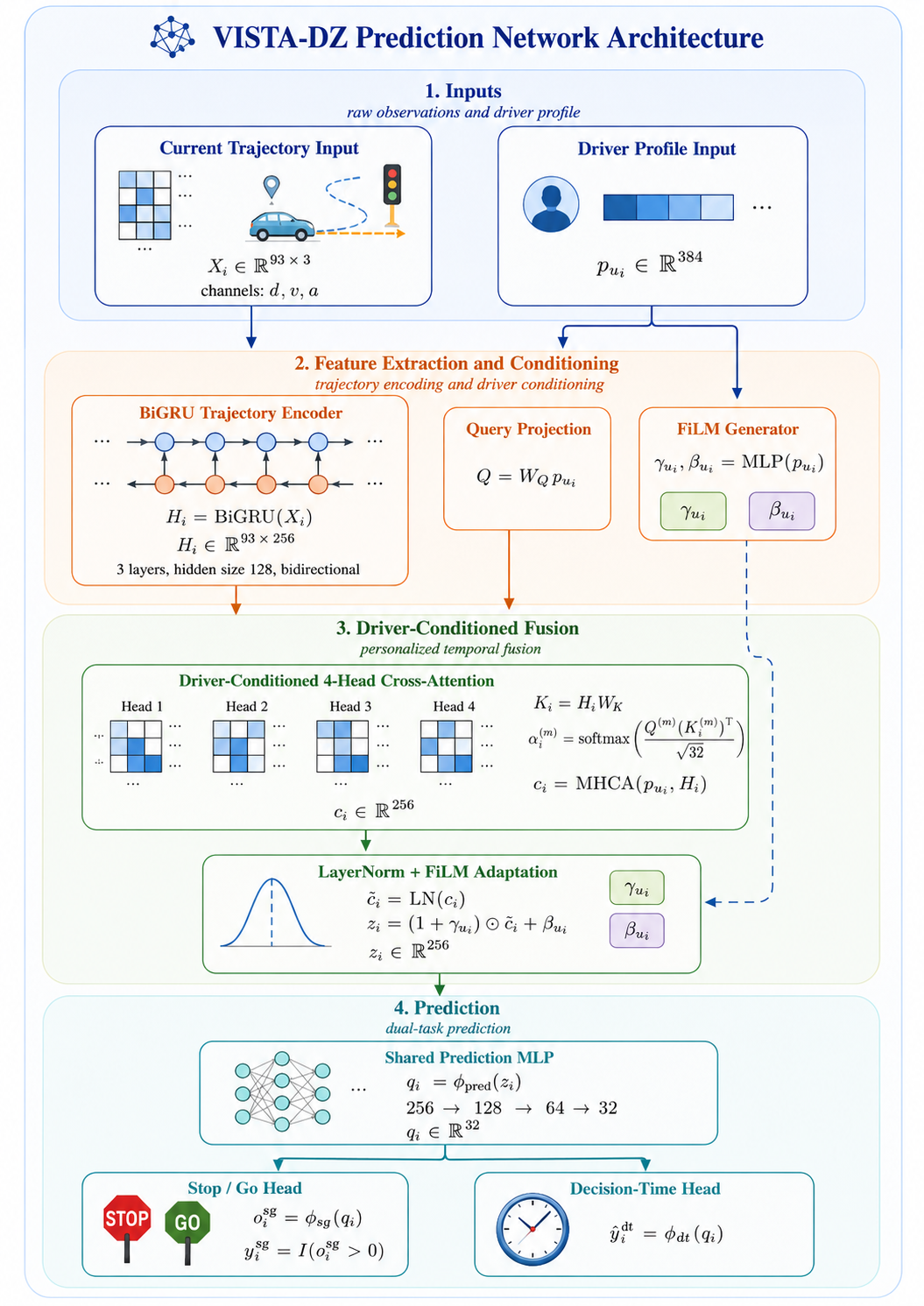}
    \caption{Architecture of the proposed VISTA-DZ prediction network. The current trajectory is encoded by a BiGRU, personalized through driver-conditioned multi-head cross-attention and FiLM modulation using the semantic driver profile, and finally passed to dual heads for stop-go classification and decision-time regression.}
    \label{fig:network_architecture}
\end{figure}

\subsubsection{BiGRU Trajectory Encoder}
\label{subsubsec:bigru_encoder}

The current trajectory sequence is encoded using a multi-layer Bidirectional GRU. Let $D_h$ denote the output dimension of the bidirectional hidden state. With hidden size $h=128$, the forward and backward hidden states are concatenated, giving $D_h=2h=256$. The encoded trajectory representation is
\begin{equation}
\mathbf{H}_i = \mathrm{BiGRU}\left(\mathbf{X}_i\right), 
\qquad 
\mathbf{H}_i \in \mathbb{R}^{T \times D_h}.
\end{equation}
In the proposed model, the BiGRU uses three recurrent layers with dropout set to 0.2 between layers. Since $T=93$ and $D_h=256$, the encoder output is $\mathbf{H}_i \in \mathbb{R}^{93 \times 256}$. Each hidden state $\mathbf{h}_{i,t} \in \mathbb{R}^{256}$ contains temporal information from both past and future directions within the pre-yellow observation window.

The use of a recurrent trajectory encoder is motivated by the sequential nature of dilemma zone decision making. A driver's decision is not determined only by the instantaneous speed or distance at a single timestep, but also by how these signals evolve before the yellow onset. For example, gradual early deceleration and late strong braking may lead to different decision interpretations even when the vehicle has a similar speed near the end of the observation window.

\subsubsection{Driver-Conditioned Multi-Head Cross-Attention}
\label{subsubsec:mhca}

After trajectory encoding, VISTA-DZ uses the semantic driver profile to select driver-relevant temporal evidence from the BiGRU hidden states. The driver profile serves as the query, while the trajectory hidden states provide the keys and values. This design allows the model to attend to different temporal portions of the same trajectory depending on the driver's historical behavioral tendency.

Let $M=4$ denote the number of attention heads and let $D_a=32$ denote the attention dimension of each head. The trajectory keys and driver query are computed as
\begin{equation}
\mathbf{K}_i = \mathrm{reshape}\left(\mathbf{H}_i \mathbf{W}_K + \mathbf{b}_K\right) 
\in \mathbb{R}^{M \times T \times D_a},
\end{equation}
\begin{equation}
\mathbf{Q}_i = \mathrm{reshape}\left(\mathbf{p}_{u_i} \mathbf{W}_Q + \mathbf{b}_Q\right) 
\in \mathbb{R}^{M \times 1 \times D_a},
\end{equation}
where $\mathbf{W}_K \in \mathbb{R}^{D_h \times (M D_a)}$ and $\mathbf{W}_Q \in \mathbb{R}^{384 \times (M D_a)}$ are learnable projection matrices. The value representation is taken as the original BiGRU hidden states:
\begin{equation}
\mathbf{V}_i = \mathbf{H}_i.
\end{equation}

For the $m$-th attention head, the attention weights over the $T$ trajectory timesteps are computed as
\begin{equation}
\boldsymbol{\alpha}_i^{(m)} =
\mathrm{softmax}\left(
\frac{
\mathbf{Q}_i^{(m)}
\left(\mathbf{K}_i^{(m)}\right)^\top
}
{\sqrt{D_a}}
\right),
\qquad
\boldsymbol{\alpha}_i^{(m)} \in \mathbb{R}^{1 \times T}.
\end{equation}
The context vector for this head is obtained by applying the attention weights to the full BiGRU hidden states:
\begin{equation}
\mathbf{c}_i^{(m)}
=
\boldsymbol{\alpha}_i^{(m)}\mathbf{H}_i,
\qquad
\mathbf{c}_i^{(m)} \in \mathbb{R}^{D_h}.
\end{equation}
Although each head computes its attention score in a lower-dimensional attention subspace, the weighted aggregation is applied to the full 256-dimensional trajectory hidden states. This allows each head to identify a different temporal pattern while preserving the complete encoded trajectory representation.

The final driver-conditioned trajectory context is obtained by averaging the head-specific context vectors:
\begin{equation}
\mathbf{c}_i
=
\frac{1}{M}
\sum_{m=1}^{M}
\mathbf{c}_i^{(m)},
\qquad
\mathbf{c}_i \in \mathbb{R}^{D_h}.
\end{equation}

\begin{figure}
    \centering
    \includegraphics[width=0.75\linewidth]{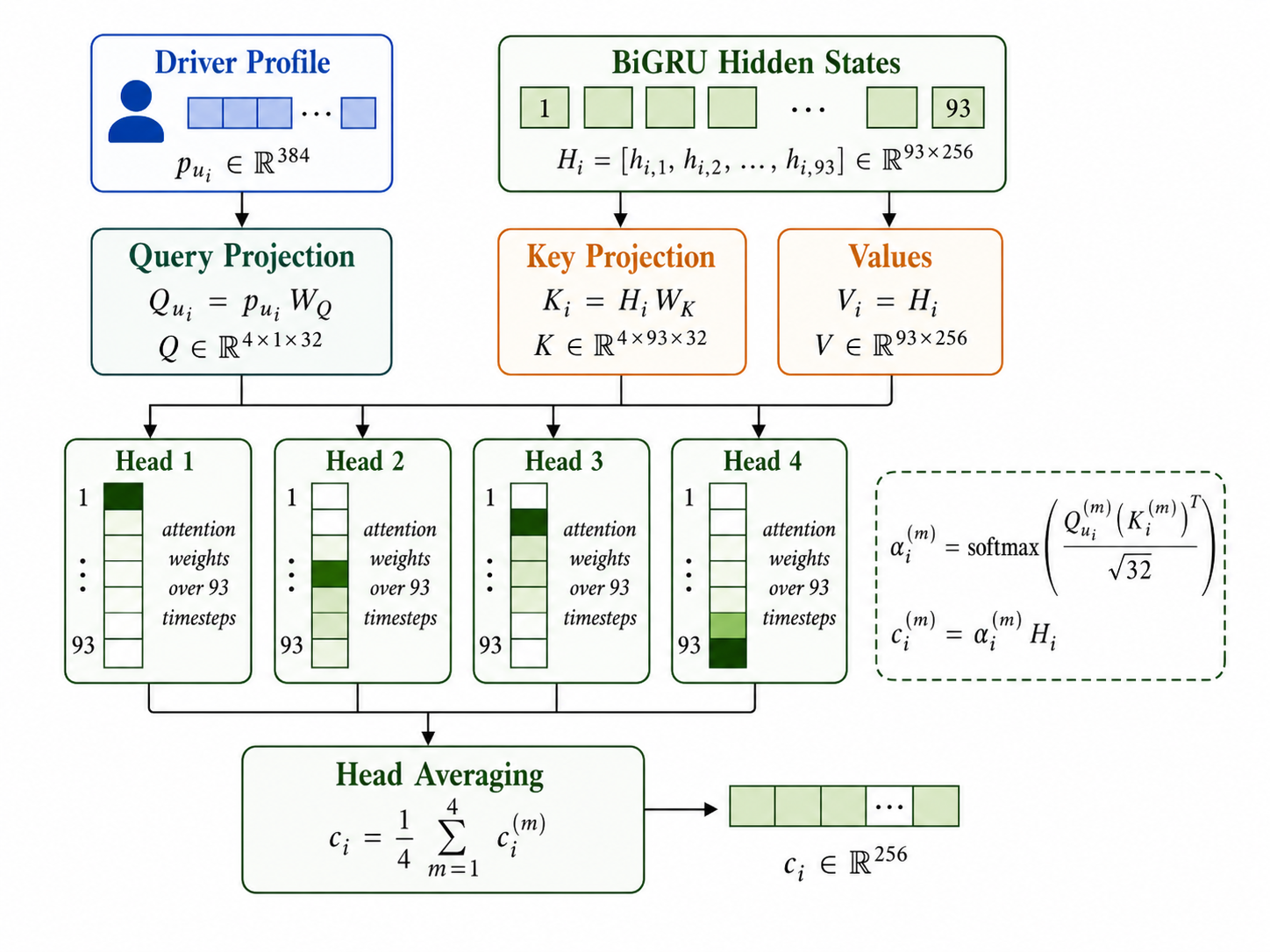}
    \caption{Detailed structure of the driver-conditioned multi-head cross-attention module. The semantic driver profile $\mathbf{p}_{u_i}$ is used as the query, while the BiGRU hidden states $\mathbf{H}_i$ provide keys and values. Four attention heads compute driver-conditioned temporal attention over 93 timesteps, and the resulting head-specific context vectors are averaged to produce the trajectory context feature $\mathbf{c}_i$.}
    \label{fig:mhca_detail}
\end{figure}

Fig.~\ref{fig:mhca_detail} illustrates the detailed structure of the driver-conditioned multi-head cross-attention module. The semantic profile $\mathbf{p}_{u_i}$ is projected as the query, while the BiGRU hidden states $\mathbf{H}_i$ are projected as keys and also used as values. The four attention heads compute distinct attention distributions over the 93 timesteps, and their context vectors are averaged to obtain $\mathbf{c}_i$. Compared with global average pooling, this design does not assume that all timesteps contribute equally. Instead, the driver profile determines which parts of the pre-yellow trajectory are most informative for the decision of that specific driver.

\subsubsection{Layer Normalization and FiLM Adaptation}
\label{subsubsec:film}

The attention context is first normalized before profile-based feature modulation:
\begin{equation}
\tilde{\mathbf{c}}_i
=
\mathrm{LayerNorm}\left(\mathbf{c}_i\right).
\end{equation}
Layer normalization stabilizes the context representation produced by multi-head attention and reduces sensitivity to large variations in the attended trajectory features.

The semantic driver profile is then mapped to a feature-wise scale vector and a feature-wise shift vector:
\begin{equation}
\boldsymbol{\gamma}_{u_i}
=
\phi_{\gamma}\left(\mathbf{p}_{u_i}\right),
\qquad
\boldsymbol{\beta}_{u_i}
=
\phi_{\beta}\left(\mathbf{p}_{u_i}\right),
\end{equation}
where $\phi_{\gamma}(\cdot)$ and $\phi_{\beta}(\cdot)$ are two independent multilayer perceptrons. Each MLP has the structure
\begin{equation}
\mathrm{Linear}(384,128)
\rightarrow
\mathrm{GELU}
\rightarrow
\mathrm{Linear}(128,256),
\end{equation}
so that
\begin{equation}
\boldsymbol{\gamma}_{u_i},
\boldsymbol{\beta}_{u_i}
\in
\mathbb{R}^{256}.
\end{equation}

The personalized feature representation is obtained by FiLM modulation:
\begin{equation}
\mathbf{z}_i
=
\left(\mathbf{1}
+
\boldsymbol{\gamma}_{u_i}\right)
\odot
\tilde{\mathbf{c}}_i
+
\boldsymbol{\beta}_{u_i},
\end{equation}
where $\odot$ denotes element-wise multiplication. The scale term $\mathbf{1}+\boldsymbol{\gamma}_{u_i}$ allows the modulation to preserve the identity mapping when the learned scale is near zero, while $\boldsymbol{\beta}_{u_i}$ shifts the feature representation according to the driver's semantic profile.

\begin{figure}
    \centering
    \includegraphics[width=0.65\linewidth]{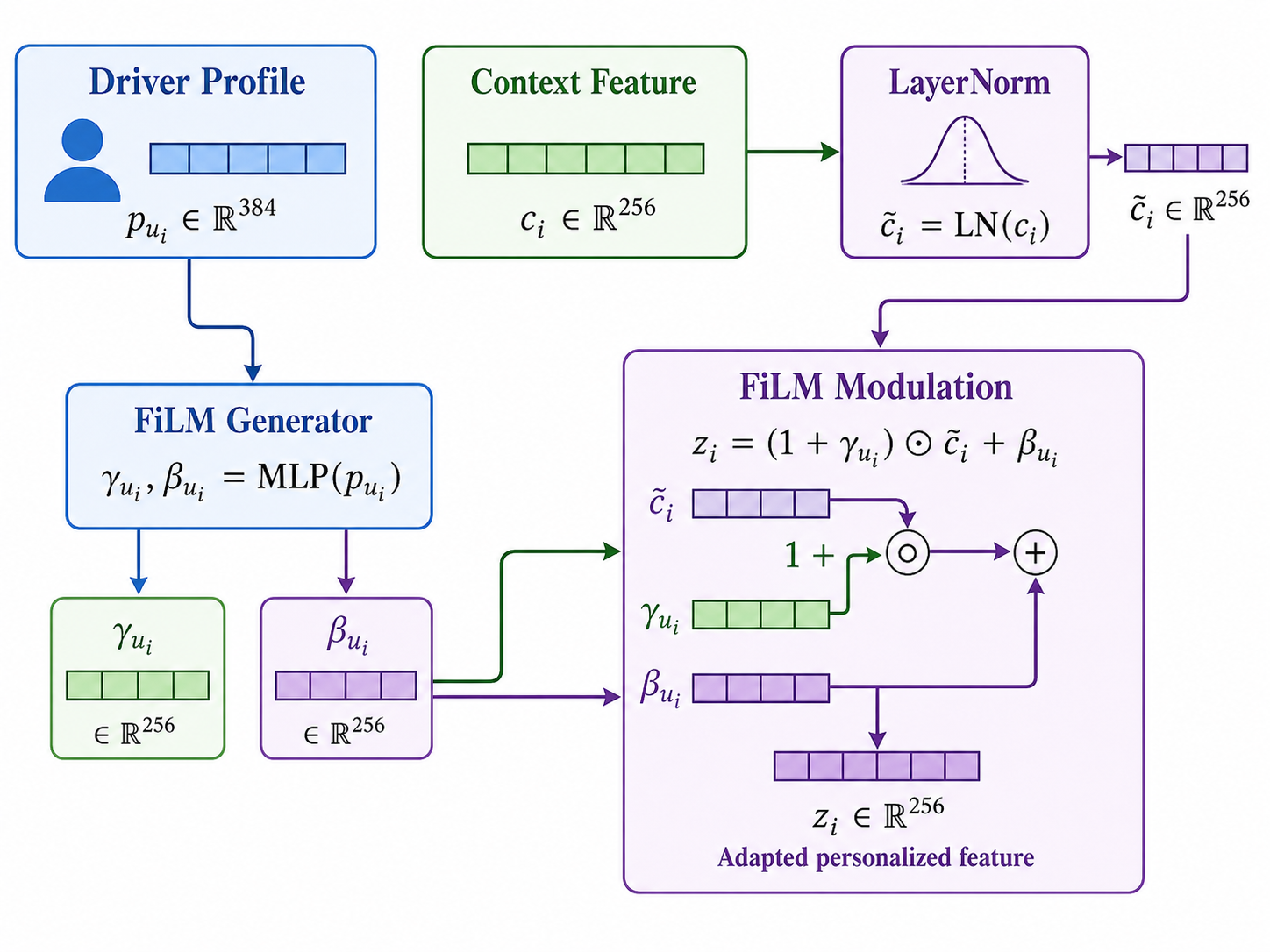}
    \caption{Detailed structure of the FiLM-based feature adaptation module. The attention context feature $\mathbf{c}_i$ is first normalized by LayerNorm. The semantic driver profile $\mathbf{p}_{u_i}$ is then mapped to feature-wise scale and shift parameters $(\boldsymbol{\gamma}_{u_i},\boldsymbol{\beta}_{u_i})$, which modulate the normalized context feature to produce the adapted personalized representation $\mathbf{z}_i$.}
    \label{fig:film_detail}
\end{figure}

Fig.~\ref{fig:film_detail} shows the detailed FiLM-based feature adaptation process. The context feature $\mathbf{c}_i$ from the attention module is first normalized. Meanwhile, the semantic driver profile $\mathbf{p}_{u_i}$ is passed through the FiLM generator to produce the feature-wise scale and shift parameters $(\boldsymbol{\gamma}_{u_i},\boldsymbol{\beta}_{u_i})$. These parameters modulate the normalized context feature and produce the adapted personalized feature $\mathbf{z}_i$. This design is well-suited for personalized dilemma zone prediction because the same observed trajectory can imply different intentions for different drivers. Rather than merely concatenating driver information to the trajectory feature, FiLM changes the interpretation of each feature dimension based on the driver's historical behavioral tendency.

\subsubsection{Dual-Task Prediction Heads}
\label{subsubsec:prediction_heads}

The personalized feature $\mathbf{z}_i$ is passed through a shared multilayer prediction trunk:
\begin{equation}
\mathbf{q}_i
=
\phi_{\mathrm{pred}}\left(\mathbf{z}_i\right),
\qquad
\mathbf{q}_i \in \mathbb{R}^{32}.
\end{equation}
The shared trunk is implemented as
\begin{equation}
\mathrm{Linear}(256,128)
\rightarrow
\mathrm{GELU}
\rightarrow
\mathrm{Dropout}(0.2)
\end{equation}
\begin{equation}
\rightarrow
\mathrm{Linear}(128,64)
\rightarrow
\mathrm{GELU}
\rightarrow
\mathrm{Dropout}(0.1)
\end{equation}
\begin{equation}
\rightarrow
\mathrm{Linear}(64,32)
\rightarrow
\mathrm{GELU}.
\end{equation}
Two task-specific linear heads are then used for classification and regression:
\begin{equation}
\hat{o}_i^{sg}
=
\phi_{sg}\left(\mathbf{q}_i\right),
\qquad
\hat{y}_i^{dt}
=
\phi_{dt}\left(\mathbf{q}_i\right),
\end{equation}
where $\hat{o}_i^{sg}$ is the raw stop-go logit and $\hat{y}_i^{dt}$ is the predicted decision time. The stop-go class is obtained by thresholding the raw logit:
\begin{equation}
\hat{y}_i^{sg}
=
\mathbb{I}
\left(
\hat{o}_i^{sg} > 0
\right).
\end{equation}

The stop-go head and decision-time head share the personalized representation but optimize different prediction objectives. This dual-task design encourages the model to learn a representation that captures both the final decision outcome and the temporal dynamics of decision commitment.

\subsection{Training Objective}
\label{subsec:training_objective}

VISTA-DZ is trained with a joint objective that combines stop-go classification and decision-time regression:
\begin{equation}
\mathcal{L} = \mathcal{L}_{sg} + \lambda_{dt}\mathcal{L}_{dt},
\end{equation}
where $\lambda_{dt}=0.2$ controls the contribution of the decision-time regression term.

\subsubsection{Stop-Go Classification Loss}
\label{subsubsec:classification_loss}

The stop-go branch is optimized using focal loss on the raw logits. Let
\begin{equation}
\hat{p}_i = \sigma\left(\hat{o}_i^{sg}\right)
\end{equation}
denote the predicted probability of the go class. For the ground-truth label $y_i^{sg}$, define
\begin{equation}
p_{t,i} = \hat{p}_i y_i^{sg} + \left(1-\hat{p}_i\right)\left(1-y_i^{sg}\right).
\end{equation}
The binary cross-entropy term is
\begin{equation}
\mathrm{BCE}_i = -y_i^{sg}\log\left(\hat{p}_i\right) - \left(1-y_i^{sg}\right)\log\left(1-\hat{p}_i\right).
\end{equation}
The focal classification loss is then
\begin{equation}
\mathcal{L}_{sg} = \frac{1}{N}\sum_{i=1}^{N}\alpha_{t,i}\left(1-p_{t,i}\right)^\gamma \mathrm{BCE}_i,
\end{equation}
where $\gamma=2.0$ is the focusing parameter and $\alpha=0.5$ is used for class balancing. This loss down-weights easily classified samples and encourages the model to focus on more ambiguous cases near the stop-go boundary.

\subsubsection{Decision-Time Regression Loss}
\label{subsubsec:regression_loss}

The decision-time branch is optimized using mean squared error:
\begin{equation}
\mathcal{L}_{dt} = \frac{1}{N}\sum_{i=1}^{N}\left(\hat{y}_i^{dt} - y_i^{dt}\right)^2.
\end{equation}
The regression loss is weighted by $\lambda_{dt}=0.2$ because decision-time estimation is treated as a complementary task to stop-go prediction. This prevents the regression term from dominating the optimization while still encouraging the shared representation to encode temporal decision information.

\subsubsection{Optimization and Regularization}
\label{subsubsec:optimization}

The model is trained using AdamW with learning rate $5\times10^{-5}$ and weight decay $1\times10^{-4}$. A linear warmup schedule is applied for the first 30 epochs, increasing the learning rate from $1\times10^{-6}$ to $5\times10^{-5}$. After warmup, cosine decay is used to gradually reduce the learning rate toward $1\times10^{-6}$. Gradient clipping with maximum norm 1.0 is applied to stabilize recurrent and attention-based training.

The batch size is set to 32. During training, online Gaussian perturbation with standard deviation $\sigma=0.01$ is added to the trajectory input as data augmentation:
\begin{equation}
\tilde{\mathbf{X}}_i = \mathbf{X}_i + \boldsymbol{\epsilon}, \qquad \boldsymbol{\epsilon} \sim \mathcal{N}\left(\mathbf{0}, 0.01^2\mathbf{I}\right).
\end{equation}
This augmentation is applied only during training and is disabled during evaluation. Dropout is used in both the BiGRU and the shared prediction trunk to reduce overfitting.

Overall, VISTA-DZ learns a personalized decision model by combining trajectory-level temporal encoding, semantic driver-conditioned attention, and feature-wise adaptation. The BiGRU captures the temporal evolution of the pre-yellow trajectory, the VLM-derived profile guides the model toward driver-relevant temporal evidence, and FiLM adapts the attended trajectory representation to the driver's long-term behavioral tendency before jointly predicting the stop-go decision and decision time.

\section{Datasets}

\begin{figure}
    \centering
    \begin{subfigure}[b]{0.49\linewidth}
        \centering
        \includegraphics[width=\linewidth]{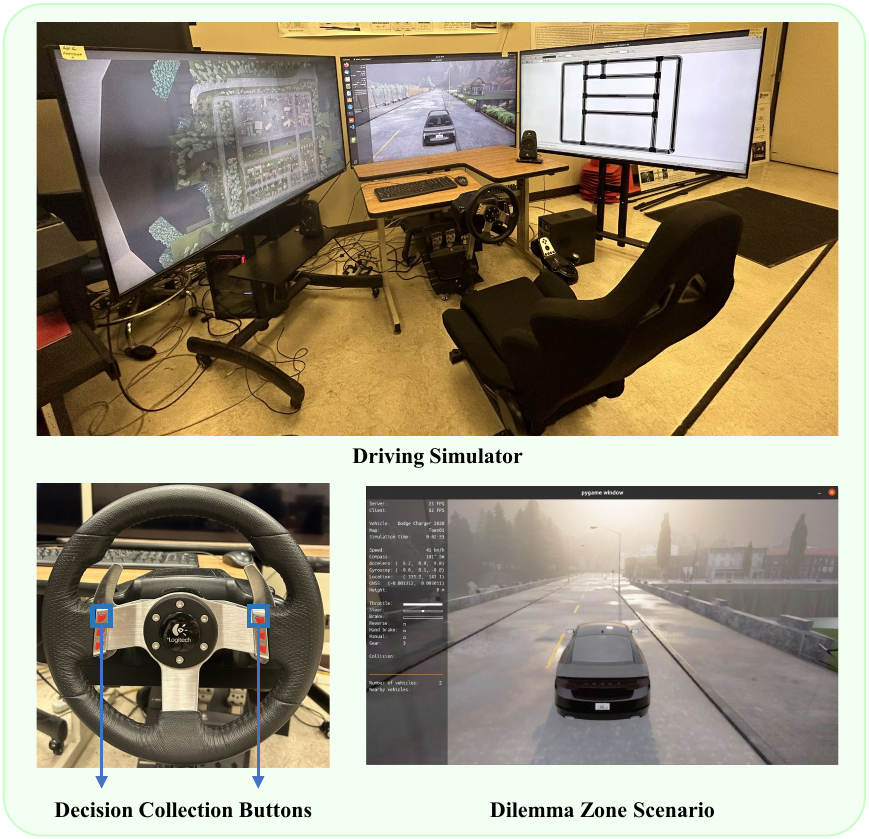}
        \caption{SDZ Dataset}
        \label{fig:sdz}
    \end{subfigure}
    \hfill
    \begin{subfigure}[b]{0.49\linewidth}
        \centering
        \includegraphics[width=\linewidth]{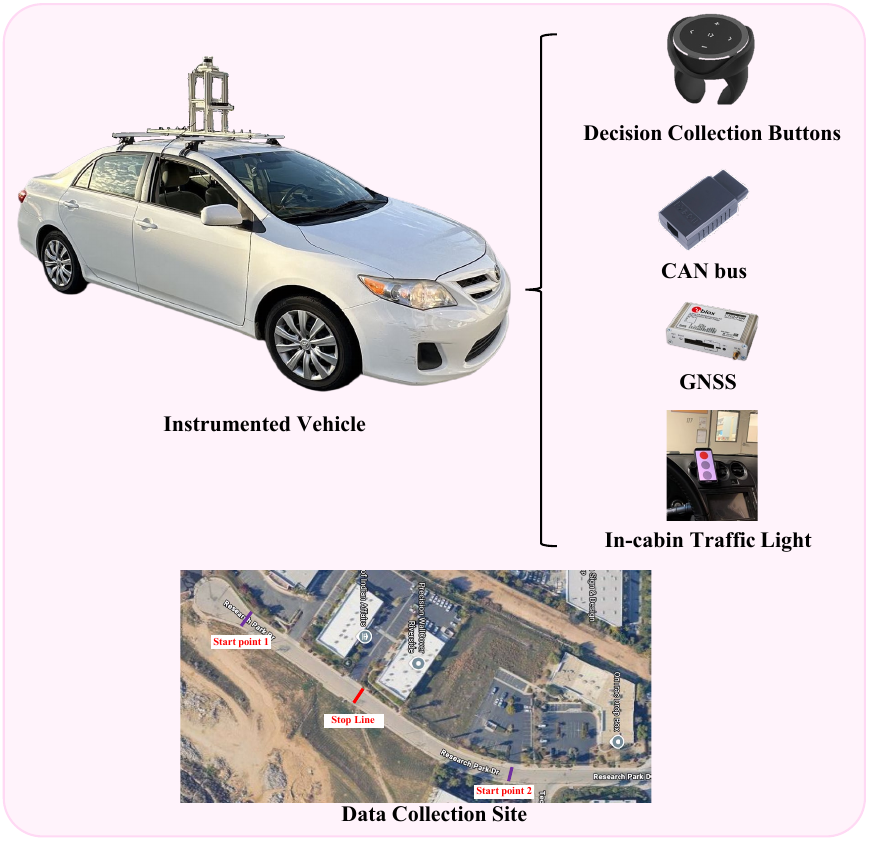}
        \caption{FDZ Dataset}
        \label{fig:fdz}
    \end{subfigure}
    \caption{Dataset Collection Setup}
    \label{fig:fdz_vehicle}
\end{figure}

This study uses two complementary datasets for evaluation. The first is the Simulator Dilemma Zone (SDZ) dataset introduced in \cite{qin2025investigating}, which provides repeated dilemma zone approach trajectories under controlled virtual traffic conditions. The second is our newly collected \textbf{Field Dilemma Zone dataset (FDZ)}, which records real-vehicle approaches under controlled field experiments designed to reproduce dilemma zone decision scenarios in a safer and more manageable setting. As described in Table~\ref{tab:datasets}, the SDZ dataset comprises 961 trajectories collected from 20 drivers, while the FDZ dataset comprises 172 trajectories from 6 drivers. As shown in Figure~\ref{fig:sdz}, the SDZ dataset was collected using the CARLA simulator with a Logitech steering wheel kit as the input device to capture driver operations; two buttons on the steering wheel were assigned to record stop–go decisions. The SDZ dataset encompasses three categories of information: complete time-series trajectory data, trajectory-level stop-go decision labels, and driver-level personalized behavior statistics. Specifically, the trajectory data record dynamic features including vehicle position, speed, acceleration, throttle input, brake input, signal state, and distance to the stop line; the decision data provide the stop-go label and the corresponding decision-state information for each trajectory; and the personalized behavior statistics summarize each driver's average decision time, speed at decision point, distance to the stop line at decision, and probability of selecting a go decision. The FDZ dataset was collected via an instrumented vehicle, as shown in Figure~\ref{fig:fdz}, which is equipped with a GNSS receiver and an IMU for trajectory retrieval. An in-cabin traffic light was additionally installed and could be toggled manually to ensure that drivers encountered yellow-light dilemma zone scenarios in the majority of trials. Vehicle states and driver operations were recorded via the CAN bus; further details of the vehicle setup are provided in \citet{wei2025pdb}. The data collection site was selected at a dead-end road in Riverside, California, USA, to ensure the safety of volunteers with respect to surrounding traffic and to avoid disrupting normal traffic operations.

\begin{table}[t]
\centering
\caption{Summary of the datasets used in this study.}
\label{tab:datasets}
\begin{tabular}{lccc}
\toprule
Dataset & Domain & Drivers & Trajectories \\
\midrule
SDZ~\citep{qin2025investigating} & Simulation & 20 & 961 \\
FDZ (ours) & Controlled field driving & 6 & 172 \\
\bottomrule
\end{tabular}
\end{table}

The two datasets serve different roles in the evaluation. SDZ provides sufficient scale for controlled comparison under random split and LODO protocols, while FDZ is used to examine whether the learned personalization mechanisms remain effective when the model is exposed to real-vehicle trajectories collected outside the simulator. This combination allows the experiments to cover both within-domain behavior modeling and cross-domain transfer.

\section{Experiments and Results}
\label{sec:experiments}

This section evaluates the proposed VISTA-DZ framework from four complementary perspectives. First, we examine whether semantic driver profiles provide a useful personalization signal compared with trajectory-only prediction and handcrafted driver descriptors. Second, we evaluate the architectural components of the final VISTA-DZ model, including model capacity, FiLM conditioning, driver-conditioned cross-attention, and multi-head attention. Third, we assess generalization to unseen drivers through the LODO evaluation on the simulation dataset. Finally, we study cross-domain transfer between simulation and real-world driving data.

\subsection{Experimental Setup}
\label{subsec:experimental_setup}

\subsubsection{Datasets and prediction task}

The prediction task contains two outputs: a binary stop-go decision and a continuous decision time $\Delta t$. Each input sample consists of a fixed-length pre-yellow trajectory window with $T=93$ time steps, corresponding to approximately three seconds before the yellow signal onset. Each time step contains three kinematic channels: distance to the stop line, speed, and longitudinal acceleration.

The simulation dataset contains 961 trajectories from 20 drivers. Unless otherwise specified, random-split experiments use an 80\%/20\% split, resulting in 768 training trajectories and 193 testing trajectories. The real-world dataset contains 172 trajectories from six drivers. Real-world driver IDs are offset from the simulation driver IDs to avoid namespace overlap when the two domains are combined. For simulation drivers, semantic driver profiles are generated from historical trajectory statistics and trajectory visualizations. For real-world drivers, proxy semantic profiles are generated from the same type of behavioral statistics and encoded using the same text encoder, so that simulation and real-world driver profiles are represented in a shared semantic embedding space.

\subsubsection{Evaluation protocols}

We use four evaluation protocols.

First, a random 80/20 split on the simulation dataset is used to measure in-domain prediction performance. This setting evaluates whether personalization improves stop-go and decision-time prediction when all drivers are represented in both training and testing sets.

Second, the LODO evaluation is conducted on the simulation dataset. In each fold, all trajectories from one driver are held out for testing, while the model is trained on the remaining 19 drivers. The held-out driver's semantic profile is available at test time, but none of that driver's trajectories are used for model training. This protocol evaluates whether the personalization mechanism can generalize to an unseen driver.

Third, we evaluate sim-to-real transfer using three real-world testing conditions. In the zero-shot setting, the model is trained on simulation data and tested on all real-world trajectories without real-world fine-tuning. In the Sim+Real LODO setting, the model is trained on all simulation drivers and five real-world drivers, and tested on the held-out real-world driver. In the Real-only LODO setting, the model is trained only on five real-world drivers and tested on the remaining real-world driver.

Fourth, we report a full $3 \times 3$ cross-domain matrix for the final architectural variants. The three training domains are simulation-only, real-world-only, and combined simulation plus real-world data. The three testing domains are the simulation test set, the held-out real-world test set, and the combined test set.

\subsubsection{Implementation details}

All final VISTA-DZ experiments use a fixed seed of 100. The proposed model is trained with AdamW using a learning rate of $5 \times 10^{-5}$ and weight decay of $1 \times 10^{-4}$. A linear warmup schedule is applied for the first 30 epochs, followed by cosine decay. The batch size is 32, and gradient clipping is applied with a maximum norm of 1.0. Additive Gaussian noise with standard deviation 0.01 is used as training-time trajectory augmentation. The total loss is
\begin{equation}
    \mathcal{L} = \mathcal{L}_{\mathrm{focal}} + 0.2 \mathcal{L}_{\mathrm{MSE}},
\end{equation}
where $\mathcal{L}_{\mathrm{focal}}$ is applied to the stop-go classification logit with $\gamma=2$ and $\alpha=0.5$, and $\mathcal{L}_{\mathrm{MSE}}$ is applied to decision-time regression.

The early semantic-profile validation experiments and the final VISTA-DZ architectural ablation serve different purposes. The former isolates the effect of driver profile representation and fusion mechanism under a lighter personalization backbone, while the latter evaluates the components of the final proposed architecture. Therefore, numerical comparisons should be made within each table rather than directly across these two groups of experiments.

\subsection{Effect of Semantic Driver Profiling and Conditioning}
\label{subsec:semantic_profile_validation}

We first evaluate whether semantic driver profiles provide a useful personalization signal. Table~\ref{tab:semantic_profile_validation} compares five personalization designs. Traj-Only uses only the current trajectory. HCF-Pers augments the trajectory representation with handcrafted driver statistics. VLM-Cat uses the semantic driver profile through direct feature concatenation. VLM+HCF combines the semantic profile with handcrafted statistics. VLM-FiLM uses the semantic profile to modulate trajectory features through FiLM conditioning.

\begin{table}[t]
\centering
\caption{Effect of driver profile representation and conditioning on the simulation random split. Accuracy is reported in percentage.}
\label{tab:semantic_profile_validation}
\begin{tabular}{lccc}
\toprule
Method & Accuracy & DT MSE & DT MAE \\
\midrule
Traj-Only & 84.14 & 0.2224 & 0.3632 \\
HCF-Pers & 85.46 & 0.2042 & 0.3471 \\
VLM-Cat & 87.22 & 0.1957 & 0.3376 \\
VLM+HCF & 85.02 & 0.2066 & 0.3403 \\
VLM-FiLM & \textbf{89.48} & \textbf{0.1902} & \textbf{0.3327} \\
\bottomrule
\end{tabular}
\end{table}

The results show a consistent benefit from driver-level personalization. HCF-Pers improves stop-go accuracy from 84.14\% to 85.46\%, indicating that historical driver behavior contains information that is not fully captured by the current pre-yellow trajectory. Replacing handcrafted statistics with semantic VLM profiles further increases accuracy to 87.22\% and reduces decision-time error, suggesting that language-based driver profiles provide a richer representation of driver tendency than a small set of scalar descriptors.

The hybrid VLM+HCF variant does not improve over VLM-Cat. Its accuracy decreases to 85.02\%, and its decision-time errors are also slightly higher than those of VLM-Cat. This result suggests that simply concatenating handcrafted statistics with semantic embeddings may introduce redundant or weakly aligned information. In contrast, VLM-FiLM achieves the best performance across all three metrics, reaching 89.48\% accuracy with the lowest MSE and MAE. The improvement from VLM-Cat to VLM-FiLM shows that the way a driver profile is injected into the prediction network is important. FiLM conditioning allows the semantic profile to rescale and shift trajectory features, which is more expressive than direct concatenation.

These findings motivate the design of the final VISTA-DZ model. Rather than treating the semantic driver profile as an additional feature vector appended to the trajectory representation, VISTA-DZ uses the profile as an active conditioning signal. The profile first guides temporal evidence selection through driver-conditioned cross-attention and then modulates the resulting trajectory context through FiLM.

\subsection{Ablation Study of the VISTA-DZ Architecture}
\label{subsec:vista_ablation}

We next evaluate the contribution of each architectural component in the final VISTA-DZ model. Table~\ref{tab:vista_ablation} compares five variants under the same simulation 80/20 split. GRU-NP is a small BiGRU without personalization. GRU-NP-L increases the recurrent backbone capacity but still does not use a driver profile. GRU+FiLM adds semantic profile-based FiLM conditioning to the small BiGRU. GRU+XA+FiLM further introduces single-head driver-conditioned cross-attention. VISTA-DZ is the full model with a larger BiGRU backbone, four-head cross-attention, LayerNorm, and FiLM modulation.

\begin{table*}[t]
\centering
\caption{Architectural ablation of VISTA-DZ on the simulation random split. Accuracy is reported in percentage.}
\label{tab:vista_ablation}
\begin{tabular}{lcccccc}
\toprule
Method & Driver profile & Backbone & Attention & Conditioning & Params & Accuracy \\
\midrule
GRU-NP & None & BiGRU-small & None & None & 111,394 & 87.05 \\
GRU-NP-L & None & BiGRU-large & None & None & 738,338 & 87.56 \\
GRU+FiLM & Semantic profile & BiGRU-small & None & FiLM & 177,314 & 91.19 \\
GRU+XA+FiLM & Semantic profile & BiGRU-small & Single-head XA & FiLM & 210,210 & 92.23 \\
VISTA-DZ & Semantic profile & BiGRU-large & 4-head MHCA & FiLM & 985,634 & \textbf{93.26} \\
\bottomrule
\end{tabular}
\end{table*}

The comparison between GRU-NP and GRU-NP-L shows that increasing model capacity alone has a limited effect. Although GRU-NP-L has more than six times the number of parameters of GRU-NP, its accuracy improves only from 87.05\% to 87.56\%. This indicates that the performance gain of the proposed framework cannot be explained simply by a larger recurrent backbone.

The largest single improvement comes from adding semantic personalization through FiLM. GRU+FiLM improves accuracy from 87.05\% to 91.19\%, corresponding to a 4.14 percentage point gain over the no-personalization baseline. This confirms the importance of driver-level semantic context for dilemma zone prediction. Adding driver-conditioned cross-attention further improves accuracy to 92.23\%. This gain indicates that the semantic profile is useful not only for feature-wise modulation but also for selecting driver-relevant temporal evidence from the approach trajectory.

The full VISTA-DZ model achieves the best accuracy of 93.26\%. Compared with GRU-NP, the full model improves accuracy by 6.21 percentage points. Compared with GRU+XA+FiLM, it adds another 1.03 percentage points, suggesting that the larger backbone and multi-head attention provide additional benefit once semantic personalization and temporal alignment are already present. Overall, the ablation results support the main design of VISTA-DZ: semantic driver profiles should be used as conditioning signals, and their interaction with trajectory features is most effective when implemented through both cross-attention and FiLM.

\subsection{Generalization to Unseen Simulation Drivers}
\label{subsec:lodo_results}

Table~\ref{tab:lodo_vista} reports the LODO results of VISTA-DZ on the simulation dataset. Each fold holds out one driver for testing and trains the model on the remaining 19 drivers. This evaluation is more stringent than the random split because the test driver's trajectories are not observed during training.

\begin{table}[t]
\centering
\caption{LODO results of VISTA-DZ on the simulation dataset. Accuracy is reported in percentage.}
\label{tab:lodo_vista}
\begin{tabular}{cccc}
\toprule
Driver & $N_{\mathrm{test}}$ & Accuracy & DT MSE \\
\midrule
0  & 52 & 92.31 & 0.1116 \\
1  & 49 & 93.88 & 0.1480 \\
2  & 51 & 92.16 & 0.0719 \\
3  & 49 & 93.88 & 0.1148 \\
4  & 46 & 91.30 & 0.2049 \\
5  & 47 & 76.60 & 0.1444 \\
6  & 38 & 92.11 & 0.4106 \\
7  & 44 & 95.45 & 0.4078 \\
8  & 51 & 86.27 & 0.1896 \\
9  & 49 & 91.84 & 0.1751 \\
10 & 49 & 93.88 & 0.1537 \\
11 & 50 & 92.00 & 0.1752 \\
12 & 45 & 95.56 & 0.1891 \\
13 & 46 & 86.96 & 0.1608 \\
14 & 47 & 100.00 & 0.1853 \\
15 & 52 & 90.38 & 0.5649 \\
16 & 48 & 93.75 & 0.1574 \\
17 & 50 & 88.00 & 0.1361 \\
18 & 51 & 92.16 & 0.0936 \\
19 & 47 & 65.96 & 0.5168 \\
\midrule
Mean $\pm$ Std & - & 90.22 $\pm$ 7.18 & 0.2156 $\pm$ 0.1371 \\
\bottomrule
\end{tabular}
\end{table}

VISTA-DZ achieves a mean LODO accuracy of 90.22\% across 20 held-out drivers, with a mean decision-time MSE of 0.2156. This result shows that the personalization mechanism can generalize to drivers whose trajectories are excluded from training, provided that a driver-level semantic profile is available at inference time. The performance remains high for most drivers, with 16 out of 20 folds achieving accuracy above 88\%.

The fold-level results also reveal substantial driver-dependent difficulty. Driver 14 reaches 100.00\% accuracy, indicating a highly consistent behavioral pattern that is easy to personalize. In contrast, Driver 19 achieves only 65.96\% accuracy and has one of the highest decision-time errors. This driver has a near-balanced stop-go tendency, making the decision boundary intrinsically ambiguous. Drivers 6, 7, 15, and 19 also show relatively high decision-time MSE, suggesting that uncertainty in commitment timing is not always aligned with classification accuracy. These results indicate that unseen-driver prediction is feasible but not uniformly easy across the driver population. The remaining errors are concentrated in drivers whose behavior is either inconsistent or close to the stop-go boundary.

\subsection{Cross-Domain Evaluation on Real-World Data}
\label{subsec:cross_domain_vista}

We further evaluate whether VISTA-DZ transfers from simulation to real-world driving data. Table~\ref{tab:cross_domain_abc} reports three real-world testing conditions. Condition A evaluates zero-shot sim-to-real transfer. Condition B evaluates real-world LODO after training on both simulation and five real-world drivers. Condition C evaluates real-world LODO using only real-world training data.

\begin{table}[t]
\centering
\caption{Cross-domain evaluation of VISTA-DZ on real-world drivers. Accuracy is reported in percentage.}
\label{tab:cross_domain_abc}
\begin{tabular}{lccc}
\toprule
Condition & Training data & Test data & Accuracy \\
\midrule
A: Sim$\rightarrow$Real zero-shot & 20 sim drivers & 6 real-world drivers & 84.88 \\
B: Sim+Real LODO & 20 sim + 5 real-world drivers & held-out real-world driver & 89.35 $\pm$ 6.90 \\
C: Real-only LODO & 5 real-world drivers & held-out real-world driver & 86.25 $\pm$ 7.49 \\
\bottomrule
\end{tabular}
\end{table}

The zero-shot sim-to-real setting achieves 84.88\% accuracy on 172 real-world trajectories. This result is encouraging because no real-world trajectories are used for training in this condition. The performance suggests that semantic driver profiles and trajectory-based conditioning can provide a transferable representation across simulation and real-world domains.

Adding real-world training data improves performance. Under Sim+Real LODO, VISTA-DZ reaches 89.35\% mean accuracy, which is 4.47 percentage points higher than zero-shot transfer. It also outperforms Real-only LODO by 3.10 percentage points. This comparison suggests that simulation data provides useful behavioral diversity when combined with a small amount of real-world data. Real-only LODO remains competitive at 86.25\%, but the smaller number of real-world training trajectories limits its ability to cover the range of driver behaviors observed at test time.

Table~\ref{tab:cross_domain_driver} gives the fold-level results for Conditions B and C. Driver 104 is the smallest test fold, with only nine trajectories, and therefore its accuracy is more sensitive to individual sample outcomes. Driver 105 has the largest test set, and the difference between Sim+Real and Real-only training is more visible for this driver. Overall, combined-domain training provides the most stable real-world generalization.

\begin{table}[t]
\centering
\caption{Fold-level real-world LODO results of VISTA-DZ. Accuracy is reported in percentage.}
\label{tab:cross_domain_driver}
\begin{tabular}{cccc}
\toprule
Held-out driver & $N_{\mathrm{test}}$ & Sim+Real LODO & Real-only LODO \\
\midrule
100 & 19 & 89.47 & 94.74 \\
101 & 18 & 100.00 & 94.44 \\
102 & 36 & 94.44 & 91.67 \\
103 & 26 & 88.46 & 80.77 \\
104 & 9  & 77.78 & 77.78 \\
105 & 64 & 85.94 & 78.12 \\
\midrule
Mean $\pm$ Std & - & 89.35 $\pm$ 6.90 & 86.25 $\pm$ 7.49 \\
\bottomrule
\end{tabular}
\end{table}

\subsection{Full Cross-Domain Matrix}
\label{subsec:full_cross_domain}

Finally, Table~\ref{tab:full_cross_domain} reports the full $3 \times 3$ cross-domain accuracy matrix for all five final architectural variants. This experiment evaluates how each model behaves under simulation-only, real-world-only, and combined-domain training.

\begin{table}[t]
\centering
\caption{Full $3 \times 3$ cross-domain stop-go accuracy for the final architectural variants. Values are reported in percentages.}
\label{tab:full_cross_domain}
\begin{tabular}{llccc}
\toprule
Method & Training domain & sim\_test & rw\_test & combined\_test \\
\midrule
GRU-NP 
& sim\_only & 84.46 & 78.49 & 81.64 \\
& rw\_only & 80.14 & 79.49 & 80.11 \\
& combined & 85.23 & 79.66 & 84.23 \\
\midrule
GRU-NP-L 
& sim\_only & 84.46 & 76.74 & 80.82 \\
& rw\_only & 77.55 & 78.85 & 77.80 \\
& combined & 85.23 & \textbf{82.39} & 84.65 \\
\midrule
GRU+FiLM 
& sim\_only & 86.01 & 78.49 & 82.47 \\
& rw\_only & 81.26 & 78.65 & 80.98 \\
& combined & 87.22 & 79.09 & 86.03 \\
\midrule
GRU+XA+FiLM 
& sim\_only & 88.60 & 76.74 & 83.01 \\
& rw\_only & 80.66 & 78.39 & 80.42 \\
& combined & 89.29 & 81.13 & 88.18 \\
\midrule
VISTA-DZ 
& sim\_only & \textbf{90.16} & 70.93 & 81.10 \\
& rw\_only & 79.97 & 79.23 & 79.83 \\
& combined & \textbf{89.81} & 80.08 & \textbf{88.57} \\
\bottomrule
\end{tabular}
\end{table}

The full cross-domain matrix provides a more nuanced view of model behavior. On the simulation test set, VISTA-DZ obtains the best performance when trained on simulation-only data and also when trained on combined data. This confirms that the proposed architecture is effective for the source domain and benefits from the richer training set under combined-domain learning.

However, VISTA-DZ is not the best model in every cross-domain cell. In particular, when trained on simulation-only data and evaluated directly on real-world trajectories, its accuracy drops to 70.93\%. This suggests that the larger personalized model is more sensitive to simulation-specific patterns when no real-world data is available for calibration. Smaller or less expressive models can sometimes transfer more conservatively under severe domain shift.

This sensitivity is substantially reduced when real-world data is included during training. Under combined-domain training, VISTA-DZ reaches 88.57\% on the combined test set, the highest overall combined-domain accuracy among all variants. GRU+XA+FiLM also performs strongly, reaching 88.18\%, which indicates that driver-conditioned cross-attention is consistently useful even with a smaller backbone. These results suggest that the full VISTA-DZ architecture is most appropriate when a limited amount of real-world data is available, while lower-capacity personalized variants may be more conservative under pure zero-shot sim-to-real transfer.

\section{Discussion}
\label{sec:discussion}

\subsection{What Semantic Personalization Adds Beyond Model Capacity}
\label{subsec:discussion_capacity}

The results indicate that the benefit of VISTA-DZ does not mainly come from increasing the size of the recurrent backbone. In the architectural ablation, enlarging the BiGRU without using driver profiles produces only a marginal gain over the smaller non-personalized model. In contrast, introducing semantic driver profiles through FiLM conditioning yields a much larger improvement. This distinction is important for dilemma zone prediction, where the same kinematic state can correspond to different decisions depending on the driver. A larger global trajectory encoder can learn a more expressive average decision boundary, but it still treats driver heterogeneity as residual variation. Semantic personalization changes the problem formulation: the model no longer predicts from trajectory alone, but from a trajectory interpreted under a driver-specific behavioral context.

The auxiliary profile-representation results provide the same evidence from another angle. Handcrafted descriptors improve over trajectory-only prediction, confirming that driver-level information is useful. However, semantic profiles outperform scalar descriptors and are most effective when used for modulation rather than direct concatenation. Scalar descriptors such as average speed, average decision distance, or go probability summarize useful tendencies, but they compress within-driver behavior into a small set of point estimates. Two drivers may have similar go probabilities while following different behavioral patterns, for example, through consistently cautious decisions or through a mixture of aggressive clearing and early stopping. A semantic profile can represent this distinction more flexibly, especially when it is derived from historical trajectory patterns and encoded in a continuous language-based embedding space.

The weaker performance of the hybrid VLM+HCF setting also deserves attention. Adding more driver-level variables does not automatically improve personalization. If scalar statistics and semantic embeddings encode overlapping or partially inconsistent priors, simple concatenation may increase representational burden rather than improve prediction. This suggests that the quality and integration of driver information matter more than the raw amount of personalized input.

\subsection{Cross-Attention and FiLM Play Different Roles}
\label{subsec:discussion_attention_film}

The improvement from GRU+FiLM to GRU+XA+FiLM and then to VISTA-DZ suggests that cross-attention and FiLM are complementary rather than redundant. Cross-attention acts as a driver-conditioned temporal selection mechanism. It determines which portions of the pre-yellow approach should receive more weight for a given driver profile. FiLM, in contrast, acts as a feature interpretation mechanism. After the temporal context has been selected, FiLM rescales and shifts the attended representation so that the same trajectory feature can carry different meanings for different drivers.

This distinction is particularly relevant near the dilemma zone boundary. A moderate deceleration at a given distance from the stop line may indicate an early commitment to stop for a cautious driver, but it may indicate unresolved behavior for a driver who often delays the decision. Direct concatenation gives the prediction head access to both trajectory and driver profile, but it leaves the network to learn this interaction only in later layers. Cross-attention and FiLM introduce the interaction earlier and more explicitly: the profile first guides the extraction of temporal evidence and then changes the feature-level interpretation of that evidence. This design is consistent with the observed improvement from average pooling to driver-conditioned attention and from direct semantic input to FiLM-based modulation.

The multi-head design further extends this mechanism. Dilemma zone behavior is not determined by a single temporal cue. Early approach speed, late deceleration, distance evolution, and acceleration variability may each be informative in different cases. A single attention head is limited to one dominant alignment pattern, whereas multiple heads can attend to different temporal aspects of the same approach. The gain of the full VISTA-DZ model over the single-head variant is modest but consistent with this interpretation.

\subsection{Driver Heterogeneity and the Boundary of Profile-Based Prediction}
\label{subsec:discussion_driver_heterogeneity}

The LODO results show that VISTA-DZ can generalize its conditioning mechanism to drivers whose trajectories are not used during training. This is a meaningful form of cross-driver generalization, but it should be interpreted carefully. The setting is not a profile-free cold start: the held-out driver's semantic profile is available at inference. Therefore, the experiment evaluates whether the model can use a new driver's profile to condition prediction, not whether it can infer an unseen driver's behavior without any historical information.

The fold-level variation also shows that personalization has a natural boundary. Drivers with stable stop-go tendencies are easier to model because their historical profiles provide a reliable prior. In contrast, drivers whose behavior is close to a balanced stop-go split remain difficult. Driver 19 is the clearest example: its low LODO accuracy and high decision-time error indicate that some errors arise not from insufficient model capacity, but from genuinely ambiguous behavior. When a driver's historical pattern is inconsistent or when the current state lies close to the behavioral decision boundary, the semantic profile cannot fully resolve the uncertainty.

This observation has practical implications. Personalized dilemma zone prediction should not be treated as a deterministic replacement for uncertainty estimation. For drivers with stable tendencies, semantic profiles can substantially sharpen the decision boundary. For drivers with variable or context-sensitive behavior, the model should ideally report calibrated uncertainty or defer to a conservative downstream control policy. Future work should therefore consider uncertainty-aware prediction heads, driver-specific confidence estimation, and profile reliability measures.

\subsection{Simulation-to-Real Transfer and the Role of Target-Domain Calibration}
\label{subsec:discussion_cross_domain}

The cross-domain results support the usefulness of simulation data, but they also show that simulation-to-real transfer is not automatic. In the zero-shot setting trained on all simulation trajectories, VISTA-DZ achieves strong real-world accuracy without real-world fine-tuning. This suggests that the pre-yellow kinematic representation and semantic driver conditioning learned from simulation capture patterns that remain relevant in the field. However, the full $3 \times 3$ cross-domain matrix gives a more nuanced picture. When the training set is restricted to the simulation split used in the matrix experiment, the sim-only VISTA-DZ model performs poorly on the real-world test set compared with its simulation accuracy. This result should not be read as a contradiction. The zero-shot condition and the full matrix use different training pools: the former trains on all simulation trajectories, while the latter preserves a simulation holdout set and trains on only the simulation training split.

The broader point is that high-capacity personalization can become sensitive to domain-specific patterns when no target-domain data are available. A model with multi-head attention and FiLM has enough flexibility to learn detailed driver-trajectory associations in simulation. Some of these associations may not transfer directly to real vehicles because real-world trajectories differ in sensing noise, vehicle control, driver adaptation, and intersection context. Lower-capacity models may sometimes transfer more conservatively under severe domain shift, but they also give up in-domain accuracy and combined-domain performance.

The most stable setting is combined-domain training. Adding a small amount of real-world data to the simulation improves real-world LODO performance over real-only training and also produces the strongest combined-domain result in the full matrix. This finding supports a practical deployment strategy: simulation can provide behavioral diversity and controlled coverage of dilemma zone scenarios, while limited real-world data can calibrate the model to field conditions. In this sense, VISTA-DZ is best viewed as a simulation-pretrained and real-data-calibrated framework, rather than a purely zero-shot simulator-to-field model.

\subsection{Implications and Limitations}
\label{subsec:discussion_limitations}

The proposed framework has several implications for personalized driver behavior modeling. First, driver profiles can be treated as active conditioning signals rather than static covariates. This is especially relevant for transportation problems where individual behavior changes the meaning of the same physical state. Second, language-based behavioral profiles offer a way to summarize historical driving patterns at a higher level than handcrafted statistics. This may be useful when the model must transfer across data sources, sensors, or collection protocols. Third, the results suggest that simulation and real-world data should not be framed as competing alternatives. Simulation offers scale and scenario control, while real-world data provides domain calibration.

Several limitations remain. The real-world dataset contains only six drivers and 172 trajectories. Although the results are promising, a larger field dataset is needed to evaluate robustness across broader driver populations, intersections, weather conditions, and traffic contexts. The current real-world profiles are also proxy profiles generated from trajectory statistics, rather than full VLM descriptions generated from rich visual trajectory summaries. This choice improves consistency and avoids relying on heavy VLM inference during deployment, but it may also limit the expressiveness of real-world profiles.

Another limitation concerns profile availability. The LODO protocol assumes that a semantic profile exists for the held-out driver. This is reasonable for applications where historical observations are available, such as fleet vehicles, repeated commuting routes, or driver-assistance systems that adapt over time. It is less suitable for a completely new driver with no prior data. A practical system would need an online profile initialization or adaptation mechanism that can update the driver representation after only a few observed approaches.

Finally, decision-time prediction remains more challenging than stop-go classification. The semantic profile captures behavioral tendency, but precise commitment timing is also strongly constrained by physical state, signal timing, and local context. A future extension could separate the two tasks more explicitly, using semantic profiles primarily for decision tendency and physical or statistical descriptors for timing estimation. Such a design may better reflect the different sources of variability behind discrete decisions and continuous response time.

\section{Conclusion}
\label{sec:conclusion}

This paper presented VISTA-DZ, a semantic-profile-conditioned framework for personalized driver behavior prediction in the dilemma zone. The model predicts both stop-go decisions and decision time from a three-second pre-yellow trajectory window. Its main idea is to use a VLM-derived driver profile not as a simple appended feature, but as a conditioning signal that guides temporal evidence selection through multi-head cross-attention and feature interpretation through FiLM modulation.

The experiments show that semantic personalization provides a clear benefit over trajectory-only prediction and handcrafted scalar descriptors. In the final architectural ablation, increasing recurrent backbone capacity alone gives only a small improvement, whereas adding semantic FiLM conditioning provides the dominant gain. Driver-conditioned cross-attention and the full multi-head VISTA-DZ architecture further improve performance, yielding the best simulation random-split accuracy. LODO evaluation further shows that the model can generalize its conditioning mechanism to drivers whose trajectories are not included during training, provided that a driver-level profile is available at inference.

The cross-domain experiments indicate that the proposed framework can transfer from simulation to real-world driving data, but also that target-domain calibration is important. Zero-shot simulation-to-real transfer is feasible, while combined simulation and real-world training gives the most reliable performance. These findings suggest a practical path for personalized dilemma zone prediction: use simulation to cover diverse behavioral scenarios, use limited real-world data for calibration, and use semantic driver profiles to adapt the interpretation of trajectory evidence to individual drivers. Future work will focus on larger field validation, online driver profile adaptation, uncertainty-aware prediction, and a more explicit separation between decision tendency modeling and decision-time estimation.

% To print the credit authorship contribution details
\printcredits

%% Loading bibliography style file
% \bibliographystyle{model1-num-names}
\bibliographystyle{cas-model2-names} 

% Loading bibliography database
\bibliography{main}

% Biography
%\bio{}
% Here goes the biography details.
%\endbio

%\bio{pic1}
% Here goes the biography details.
%\endbio

\end{document}